\documentclass[mlabstract]{jmlr}

\usepackage{longtable}

\usepackage{booktabs}
\usepackage[load-configurations=version-1]{siunitx} 

\theorembodyfont{\upshape}
\theoremheaderfont{\scshape}
\theorempostheader{:}
\theoremsep{\newline}
\usepackage{enumitem}
\usepackage{todonotes}
\jmlrvolume{}
\firstpageno{1}

\jmlryear{2023}
\jmlrworkshop{Symmetry and Geometry in Neural Representations}

\title[Are ``Hierarchical" Visual Representations Hierarchical?]{Are ``Hierarchical" Visual Representations Hierarchical?}

  \author{Ethan Shen \quad Ali Farhadi \quad Aditya Kusupati \\ \addr University of Washington \\ \Email{\{ethans03, ali, kusupati\}@cs.washington.edu}}

\begin{document}

\maketitle
\begin{abstract}
Learned visual representations often capture large amounts of semantic information for accurate downstream applications. Human understanding of the world is fundamentally grounded in hierarchy. To mimic this and further improve representation capabilities, the community has explored ``hierarchical'' visual representations that aim at modeling the underlying hierarchy of the visual world. In this work, we set out to investigate if hierarchical visual representations truly capture the human perceived hierarchy better than standard learned representations. To this end, we create HierNet, a suite of 12 datasets spanning 3 kinds of hierarchy from the BREEDs subset of ImageNet. After extensive evaluation of Hyperbolic and Matryoshka Representations across training setups, we conclude that they \textit{do not capture hierarchy} any better than the standard representations but can assist in other aspects like search efficiency and interpretability. Our benchmark and the datasets are open-sourced at \url{https://github.com/ethanlshen/HierNet}.

\end{abstract}

\section{Introduction and Motivation}
\label{sec:intro}
As humans, our understanding of the world is fundamentally grounded in hierarchy. For example, species are classified in a hierarchical manner, starting with general orders that devolve into groups of superfamilies that finally separate into specific species. Mirroring this tendency, many vision datasets also reflect hierarchies. Some hierarchies are explicit --- ImageNet is based on synsets from WordNet~\citep{miller1995wordnet} -- while others are implicit. For example, RedCaps can easily be thought of as a taxonomy of 12M image-text pairs organized under 350 subreddits and finally joined at a single root node~\citep{desai2021redcaps}.

As a result, there has been increasing interest in developing models that can not only understand image classes but also the hierarchies that they form. One strategy is to train embeddings so their cosine similarity reflects the known semantic similarity between images~\citep{barz2019hierarchy}. \citet{kusupati2022matryoshka} goes a step further and demonstrates using adaptive Matryoshka Representations (MRs) that nest information hierarchically as the number of embedding dimensions increase. 

The task of learning hierarchies has also led to an interest in hyperbolic spaces as an alternative to Euclidean spaces for the easier embedding of complex hierarchical relationships~\citep{nickel2017poincare}. However, despite the theoretical benefits of hierarchical embeddings, improvements in downstream tasks such as classification and retrieval have been marginal compared to normally learned embeddings.  

In this paper, we aim to investigate whether hierarchical embeddings are actually better at capturing the hierarchical structure of visual data. We generate \textbf{HierNet}, a collection of 12 datasets spanning 3 diverse settings with known hierarchies from the BREEDs~\citep{santurkar2020breeds} subset of ImageNet~\citep{russakovsky2015imagenet}. For each dataset, we attempt to recreate their known hierarchy by clustering image embeddings while observing for any quantitative or qualitative benefit that hierarchical embeddings offer.  In particular, we analyze the hierarchical embeddings of two models: MERU~\citep{desai2023meru}, which is a hyperbolic version of CLIP~\citep{radford2021learning}, and MR-ResNet50, a version of ResNet50 trained with Matryoshka Representations~\citep{kusupati2022matryoshka}. We chose these two models because of the availability of their non-hierarchical counterparts -- trained without any constraints by using contrastive or cross-entropy loss appropriately. 

\section{HierNet and Methodology}

\paragraph{Dataset Creation:} HierNet is composed of 12 hierarchical datasets created with BREEDs, a subset of ImageNet that conforms to both a visual and semantic hierarchy~\citep{santurkar2020breeds}. BREEDs contains nine levels, with nodes increasing in specificity with depth. For example, ``dog" is placed at the same level as ``cat" but higher than ``bloodhound". Each BREEDs dataset is organized under a single root node with $n_{sup}$ superclasses and $n_{sub}$ subclasses per superclass. Subclasses and their images are taken directly from ImageNet.

\begin{itemize}[leftmargin=*]\vspace{-2mm}
    \itemsep 0pt
    \topsep 0pt
    \parskip 2pt
    \item \textbf{3 control datasets} are sourced from level 2 root nodes \& level 5 superclasses (r2\_l5). Each dataset has $n_{sup} = 17$ and $n_{sub} = 2$, totalling 34 subclasses (1700 images each).
    \item We also generated \textbf{4 fine-grained datasets} (level 3 root, level 5 superclass -- r3\_l5). Each has $n_{sup} = 10$ and $n_{sub} = 2$ (1000 images each). 
    \item Finally, we created \textbf{5 high-variance datasets} (level 0 root, level 5 superclass -- r0\_l5), with identical sizes to the controls.
\end{itemize}

 We create three types of datasets to evaluate the potential benefits of hierarchical embeddings across a variety of use cases and granularities. Appendix~\ref{apdx:dsets} shows the superclass-subclass composition of each dataset.

\paragraph{Cluster Quality:} For each model, we cluster image  embeddings according to the number of superclasses and subclasses present in each dataset. Similar to~\citet{nguyen2023probing}, we generate clusters using agglomerative clustering with  Ward linkage, shown by~\citet{monath2021scalable} to be optimal for image embeddings. For MERU, we cluster embeddings using hyperbolic distance metric (Section~\ref{sec:meru}) and for CLIP, MR-ResNet50, and ResNet50, we use Euclidean distance metric.

We evaluate cluster quality by comparing the discovered clusters against the ground truth labels for superclasses and subclasses. Specifically, we track adjusted mutual information (AMI) and purity. Adjusted mutual information measures the mutual information between two distinct clusterings, normalized and adjusted for chance. It quantifies to what extent two clusterings overlap. Meanwhile, purity measures the average homogeneity of clusters. Both metrics fall between $[0, 1]$, where $1$ denotes high cluster quality and $0$ low cluster quality. Equations for the two metrics are included in Appendix \ref{apdx:metrics}. 

\paragraph{Optimal Transport Distance:} While a good model will discover a hierarchy that reflects the true relationship between image classes, it is also important that a hierarchy visually aligns with the ground truth. As a result, we leverage hierarchical optimal transport (HHOT) to measure the visual difference between discovered clusterings and the ground truth hierarchy. HHOT was originally introduced by~\citet{yeaton2022hierarchical} as a method to measure the visual difference between histopathology datasets that are partitioned into distinct image slides. The technique can be easily extended to visually compare clusters, which have a similar connotation to slides.

\section{Hyperbolic Representations for CLIP}
\label{sec:meru}
\begin{figure}[h!]
        \centering
          
          {%
              \includegraphics[width=0.35\linewidth]{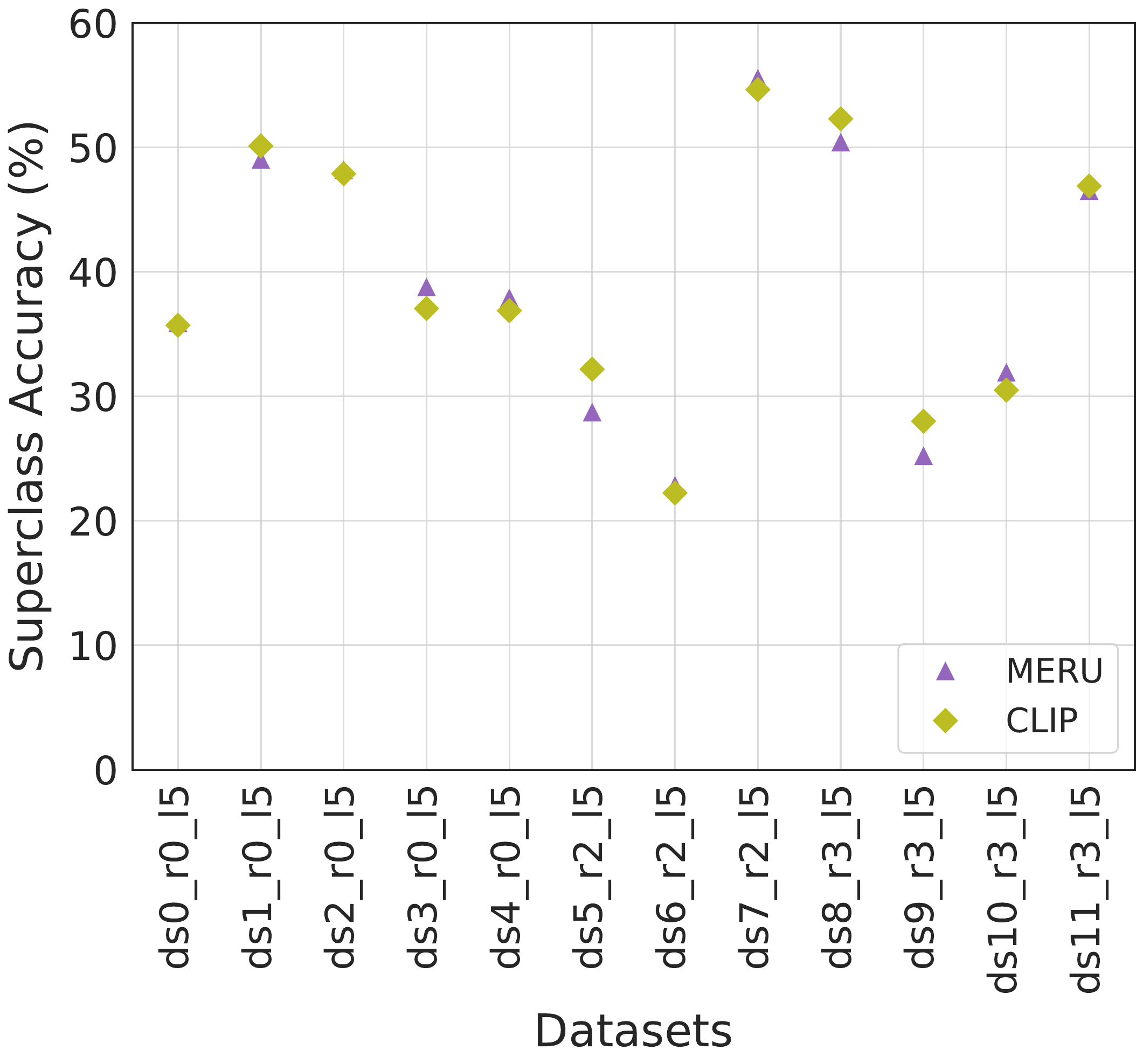}
            \qquad
              \includegraphics[width=0.35\linewidth]{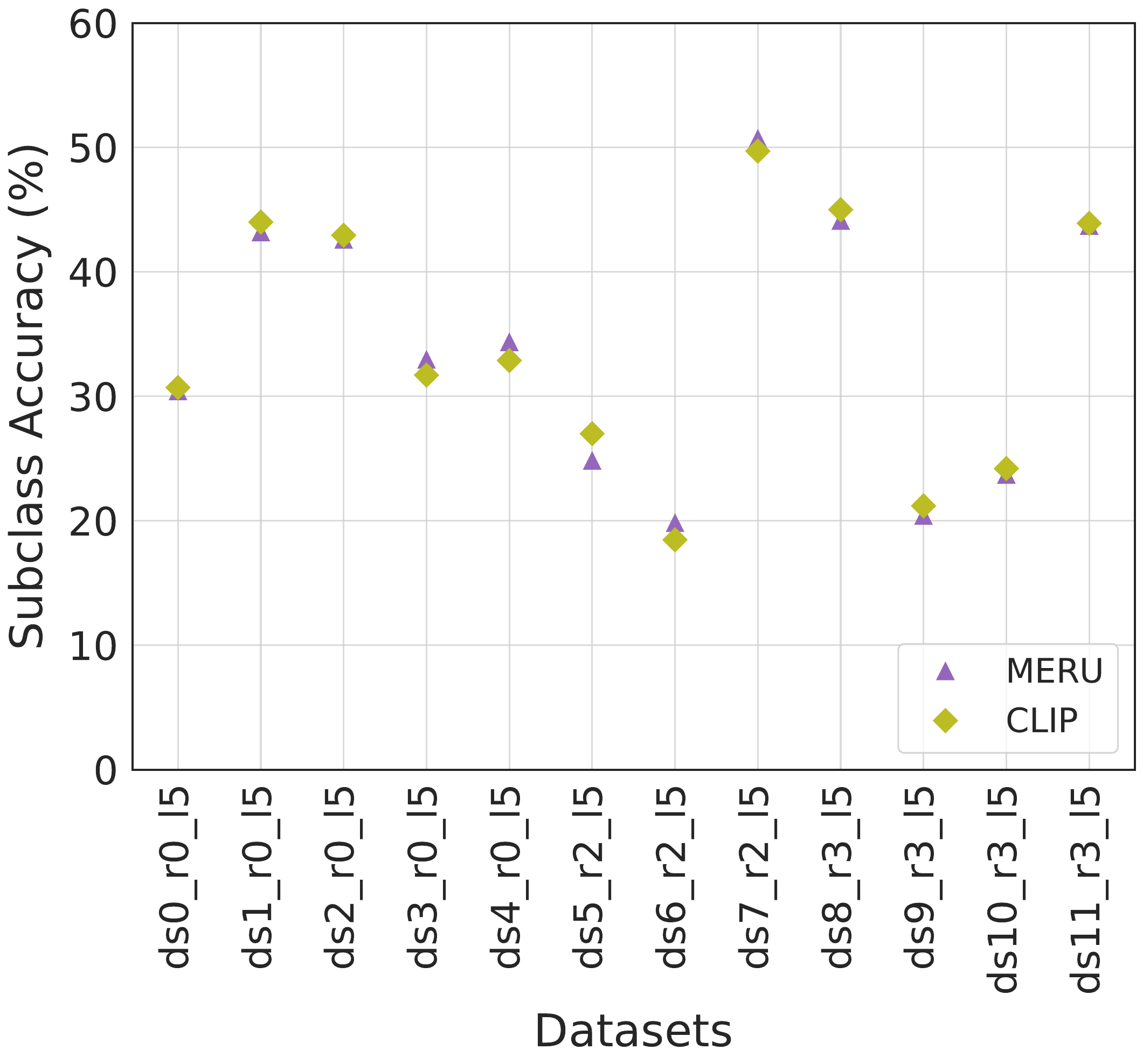}
          }
          \caption{MERU and CLIP have similar accuracies across all the datasets. A r0\_l5 denotes high-variance datasets, r2\_l5 denotes control datasets, and r3\_l5 denotes fine-grained datasets.}
          \label{fig:hyp1}
          \vspace{-6mm}
\end{figure}
\begin{figure}[h!]
    \centering
      {%
          \includegraphics[width=0.22\linewidth]{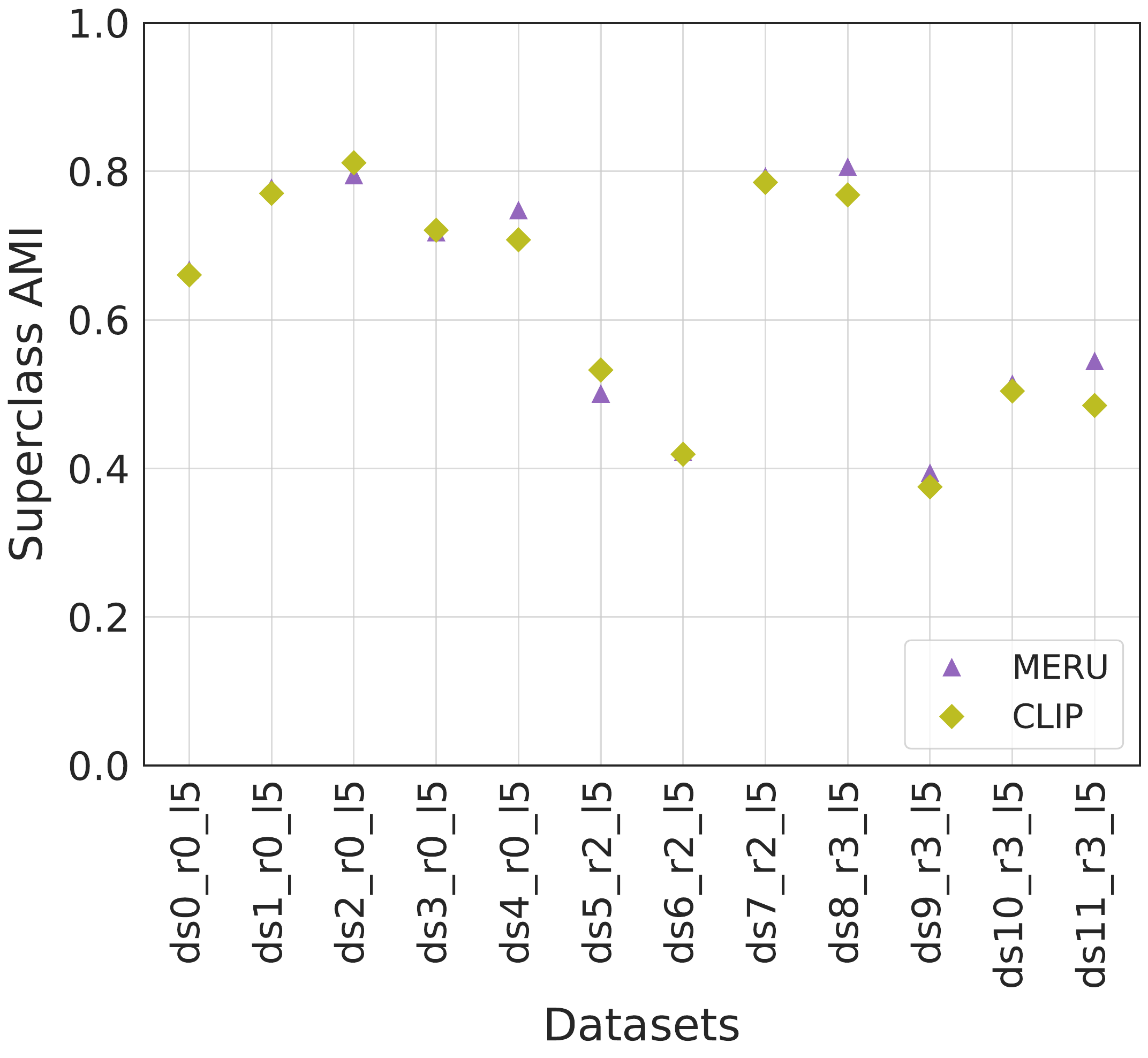}
          \includegraphics[width=0.22\linewidth]{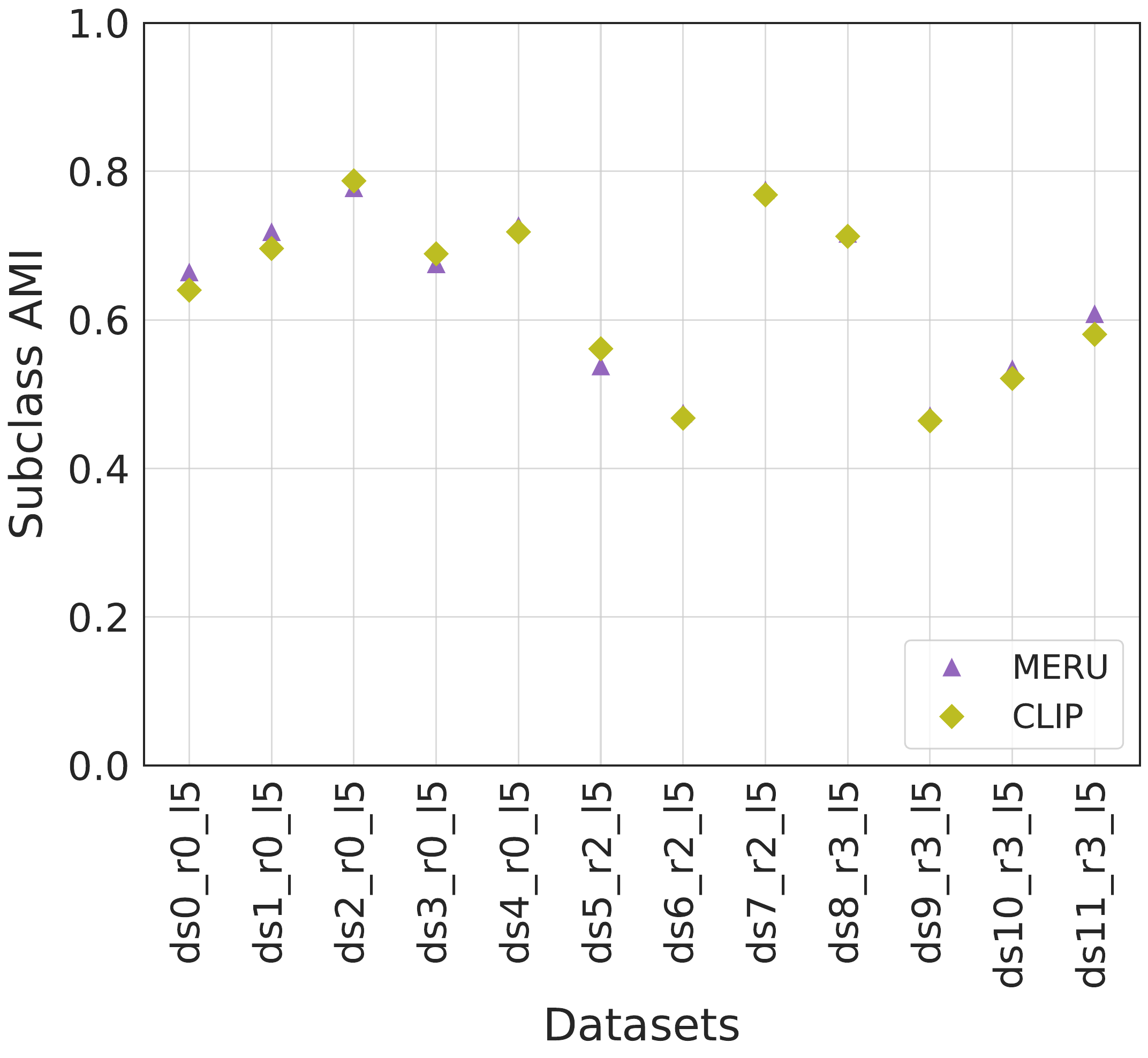}
          \includegraphics[width=0.22\linewidth]{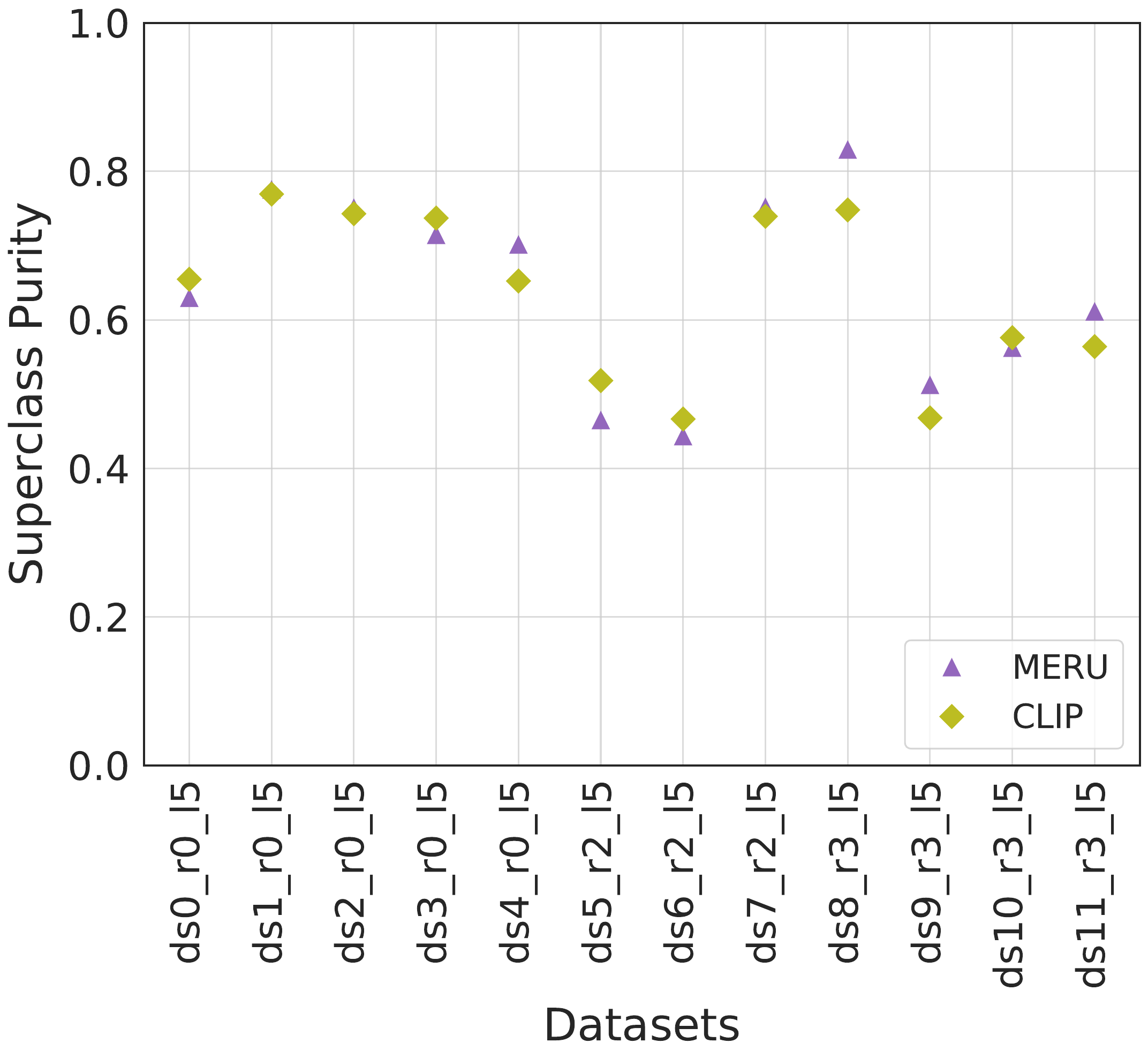}
          \includegraphics[width=0.22\linewidth]{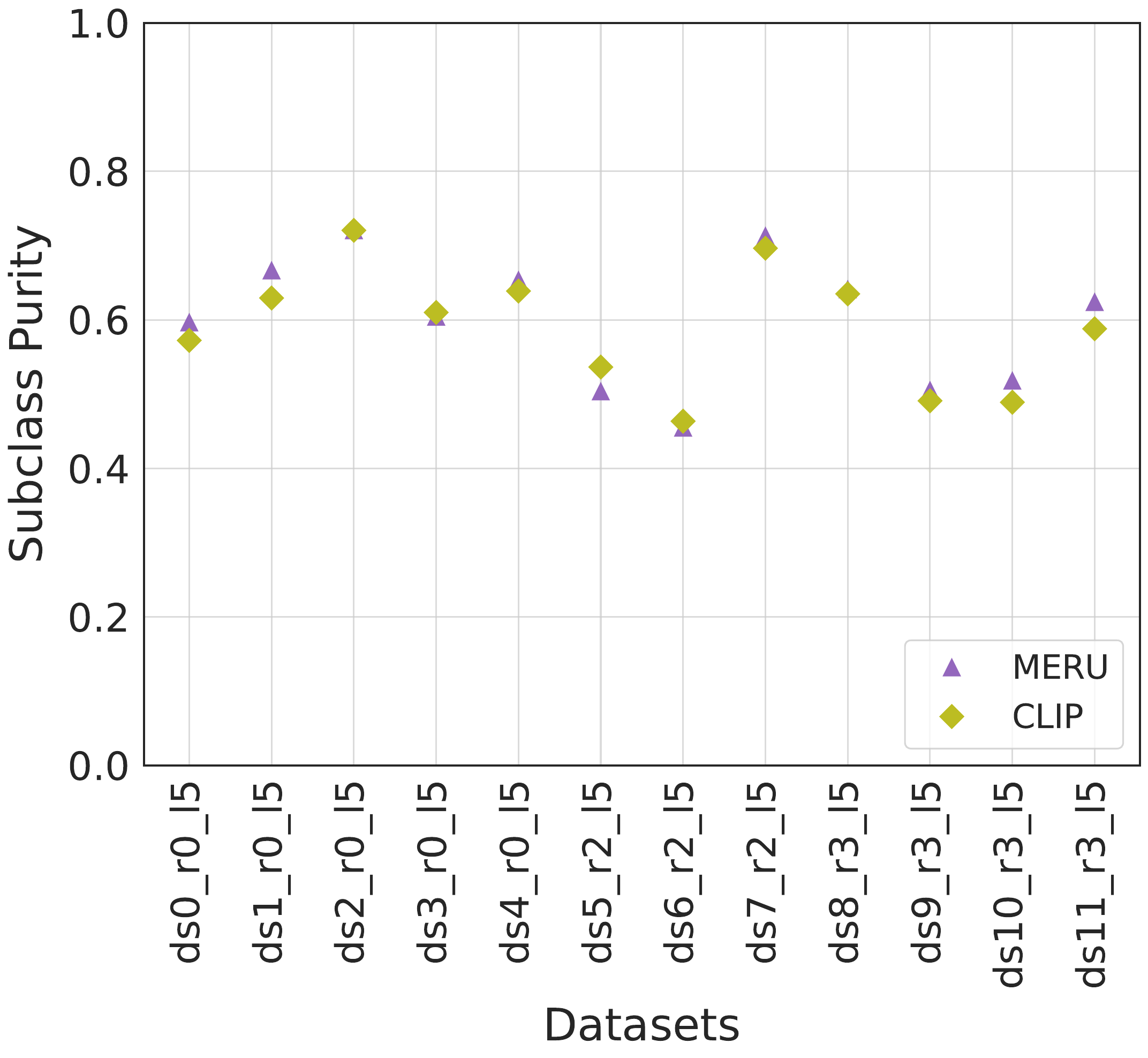}
      }
      \caption{MERU \& CLIP have a negligible difference in cluster quality across all 12 datasets.}
      \label{fig:hyp2}
      \vspace{-6mm}
\end{figure}
    \begin{figure}[htbp]
       \centering

      {%
          \includegraphics[width=0.35\linewidth]{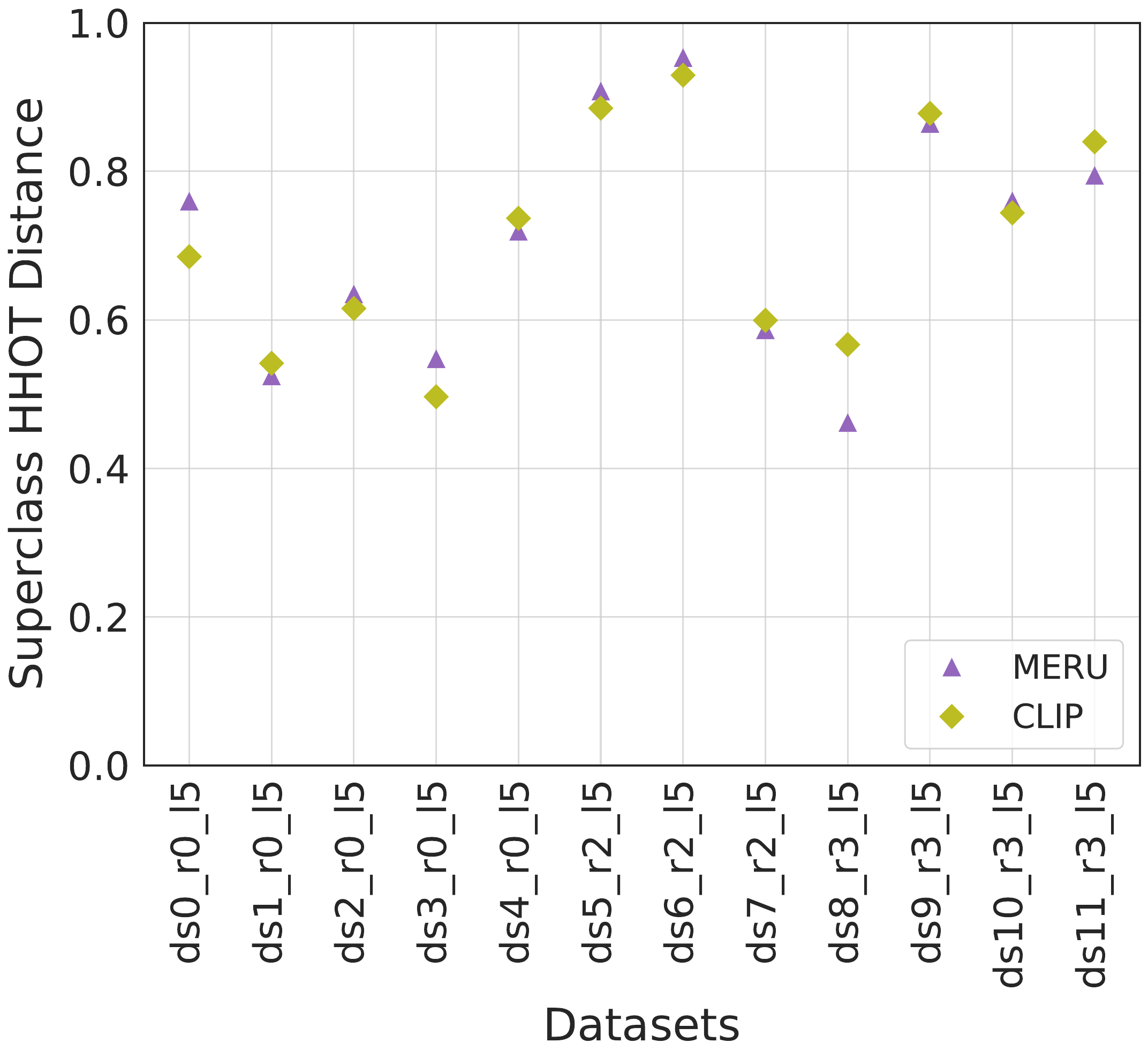}\hspace{8mm}
          \includegraphics[width=0.35\linewidth]{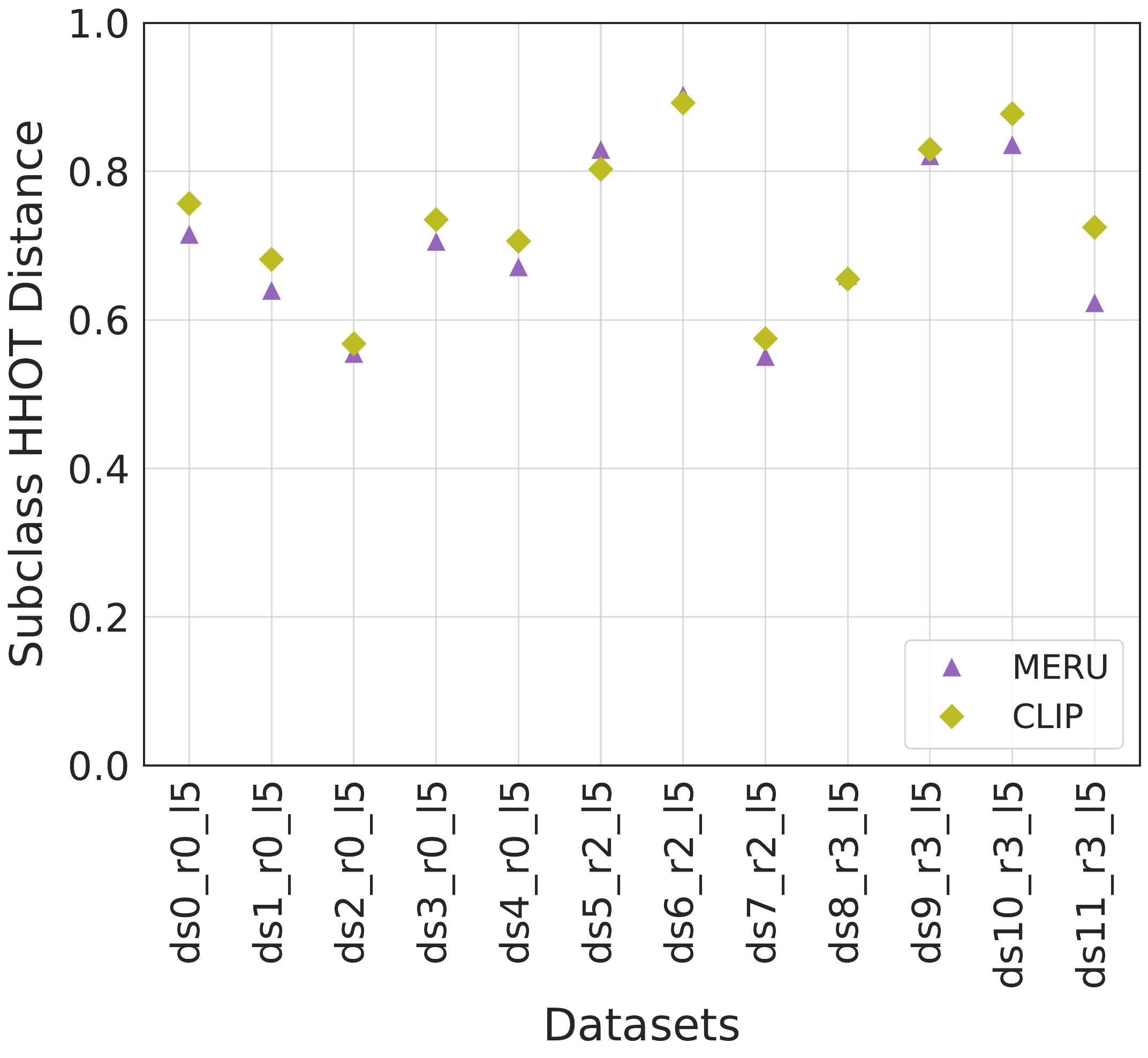}
      }
        \caption{MERU and CLIP have close HHOT distances for all datasets.}
        \label{fig:hyp3}
         \vspace{-2mm}
\end{figure}
To explore the utility of hyperbolic embeddings, we evaluate MERU against CLIP using the models from~\citet{desai2023meru}. For both superclass and subclass accuracy, MERU and CLIP perform equally on nearly all datasets (Figure~\ref{fig:hyp1}). Similarly, MERU fails to demonstrate any improvement at forming superclass and subclass clusters over CLIP, suggesting that MERU offers little actual benefit towards reconstructing hierarchical relationships. Figure \ref{fig:hyp3} supports this conclusion. HHOT distances between MERU and CLIP are almost indistinguishable. Our experiments show that hyperbolic embeddings, although theoretically sound, do not necessarily improve the retention of hierarchical information.

\section{Matryoshka Representations for ResNet50}
We also compare the performance of MR-ResNet50 embeddings against those of fixed capacity representations (FF) and PCA-reduced 2048-d FF representations. While FF embeddings provide a non-hierarchical baseline for each dimension and are retrained from scratch, PCA-reduced embeddings are a better comparison when considering compute requirements. Indeed, FF embeddings require separately trained models for each embedding dimension. On the other hand, PCA-reduced embeddings closely mimic the behavior of MR as just an extra step on a single model. We analyze the differences from 8 dimensional to 2048 dimensional representations.

    \begin{figure}[h!]
    \centering
    \vspace{-2mm}
      {%
          \includegraphics[width=0.22\linewidth]{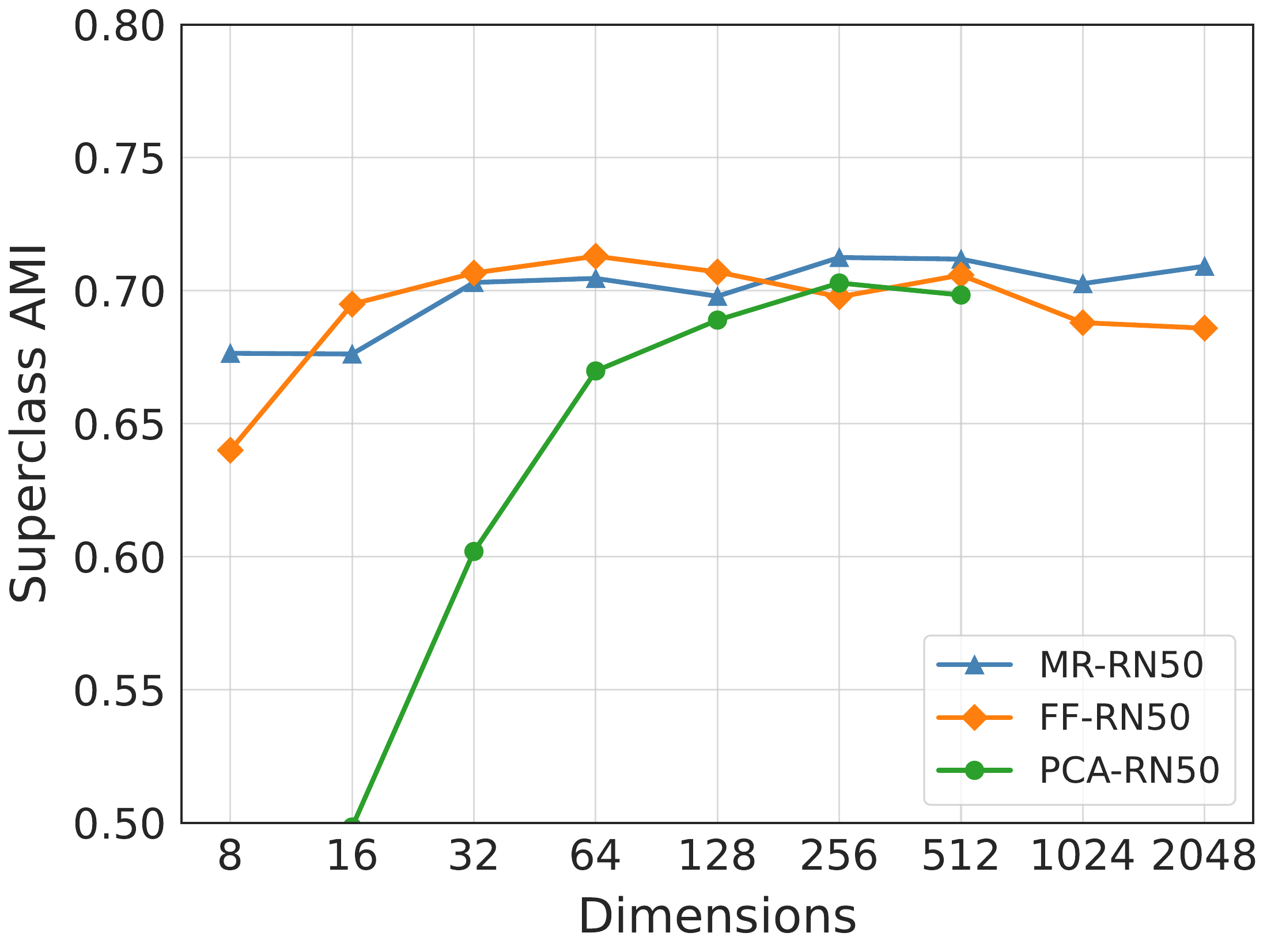}
          \includegraphics[width=0.22\linewidth]{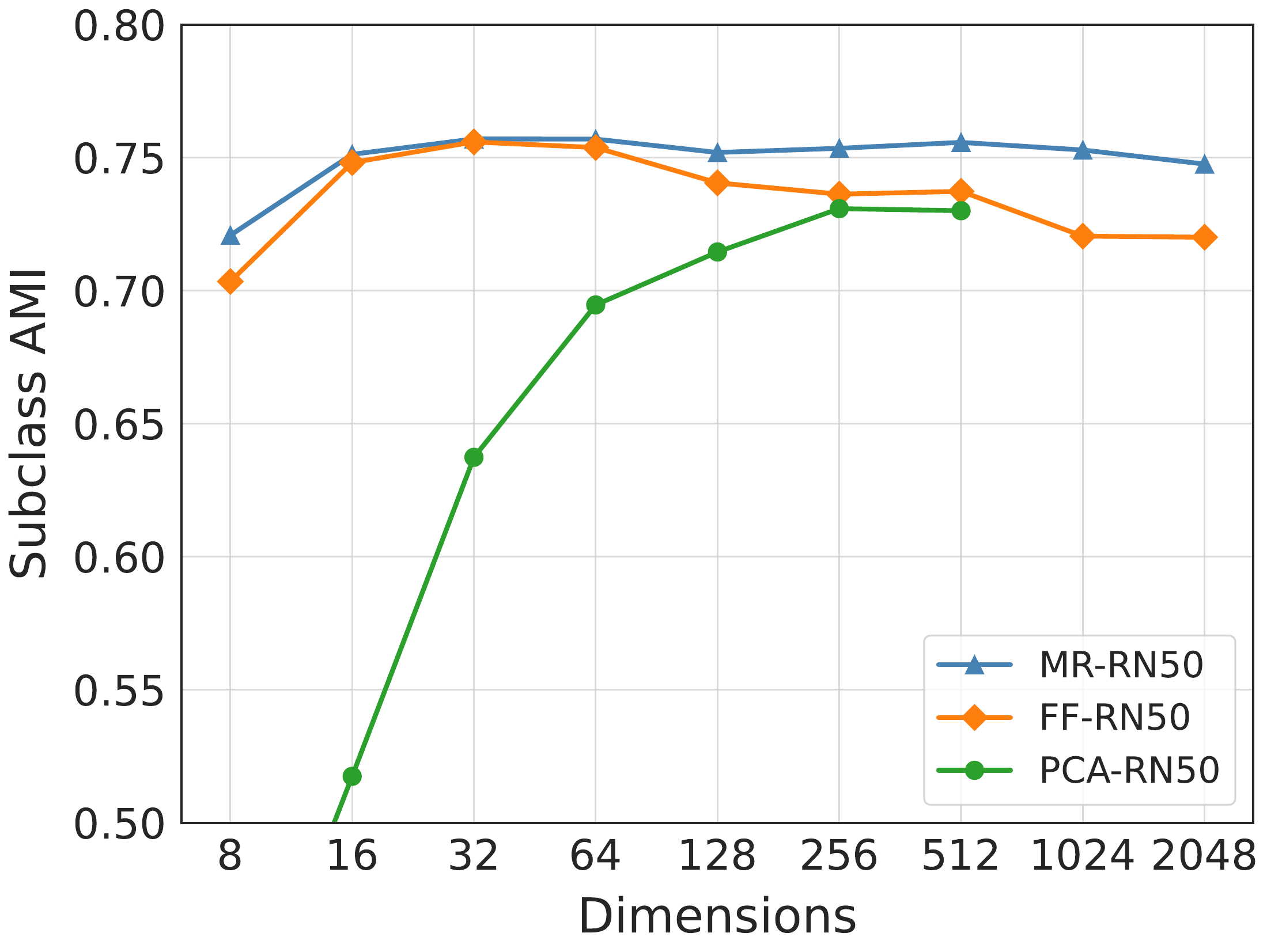}
          \includegraphics[width=0.22\linewidth]{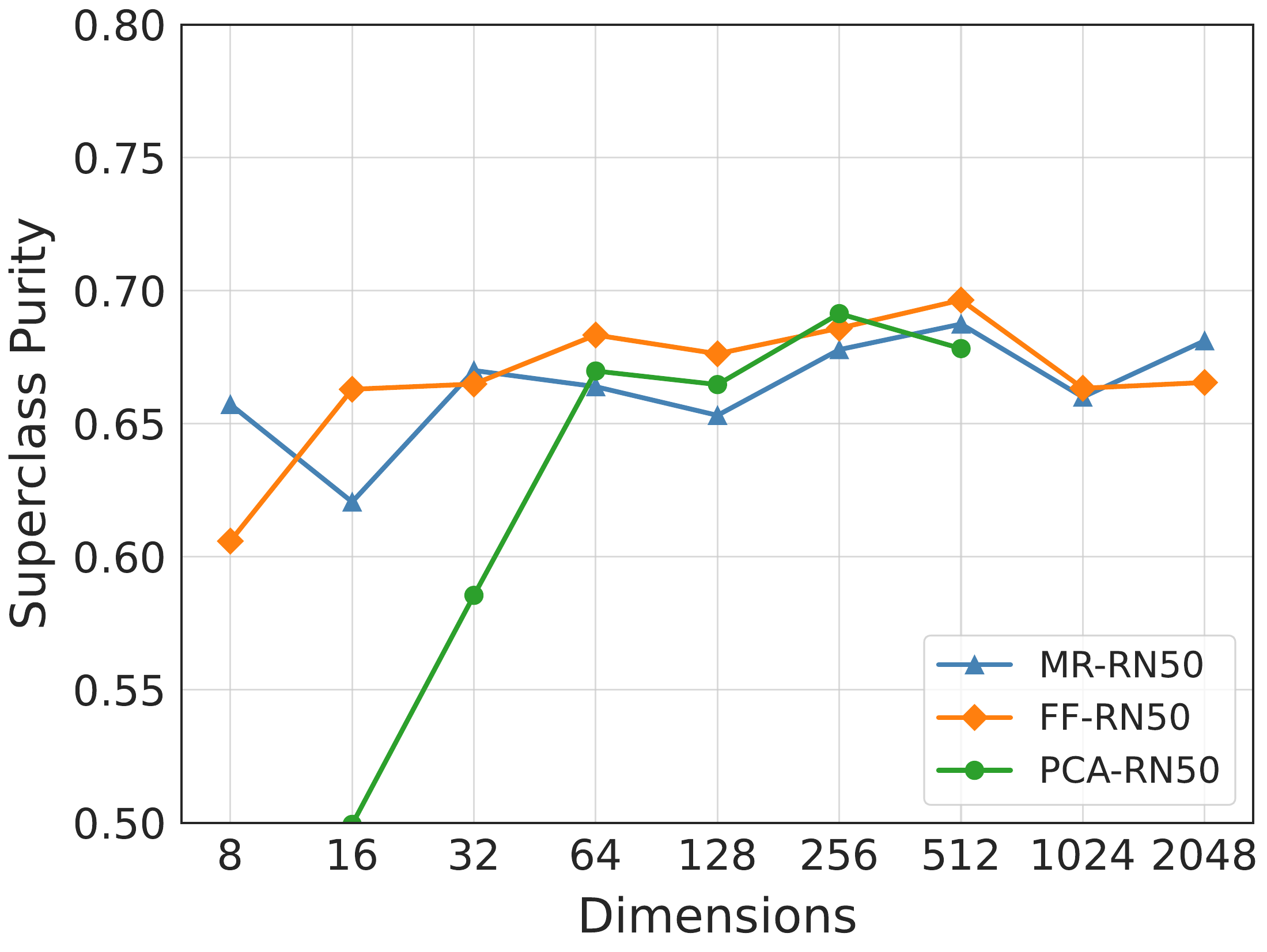}
          \includegraphics[width=0.22\linewidth]{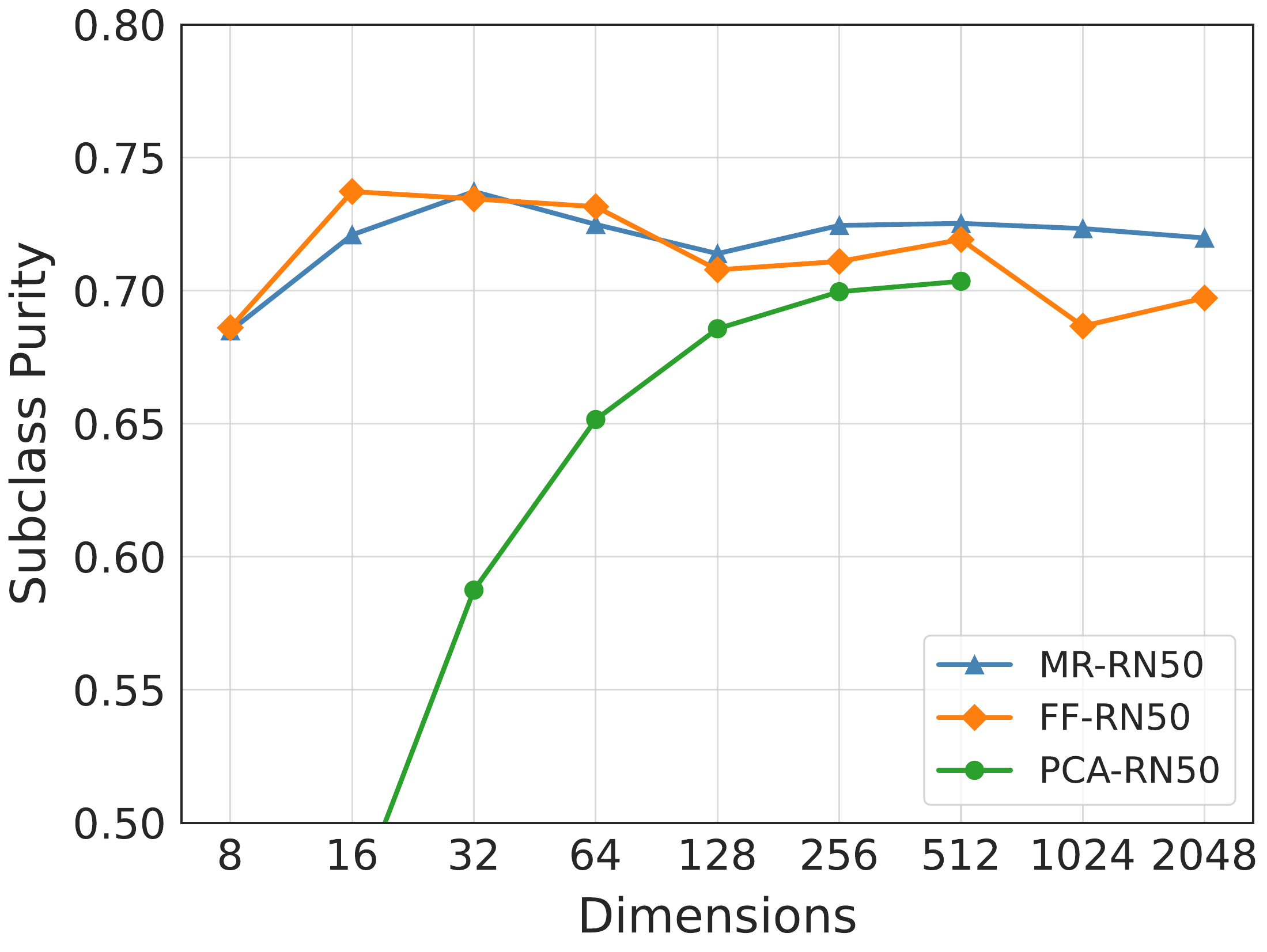}
      }
      \vspace{-2mm}
      \caption{On control datasets, average AMI and purity demonstrate that after 64-d, MR is more capable at clustering subclasses than FF or PCA representations.}
      \label{fig:c2}
      \vspace{-6mm}
    \end{figure}
\begin{figure}[h!]
       \centering

      {%
          \includegraphics[width=0.3\linewidth]{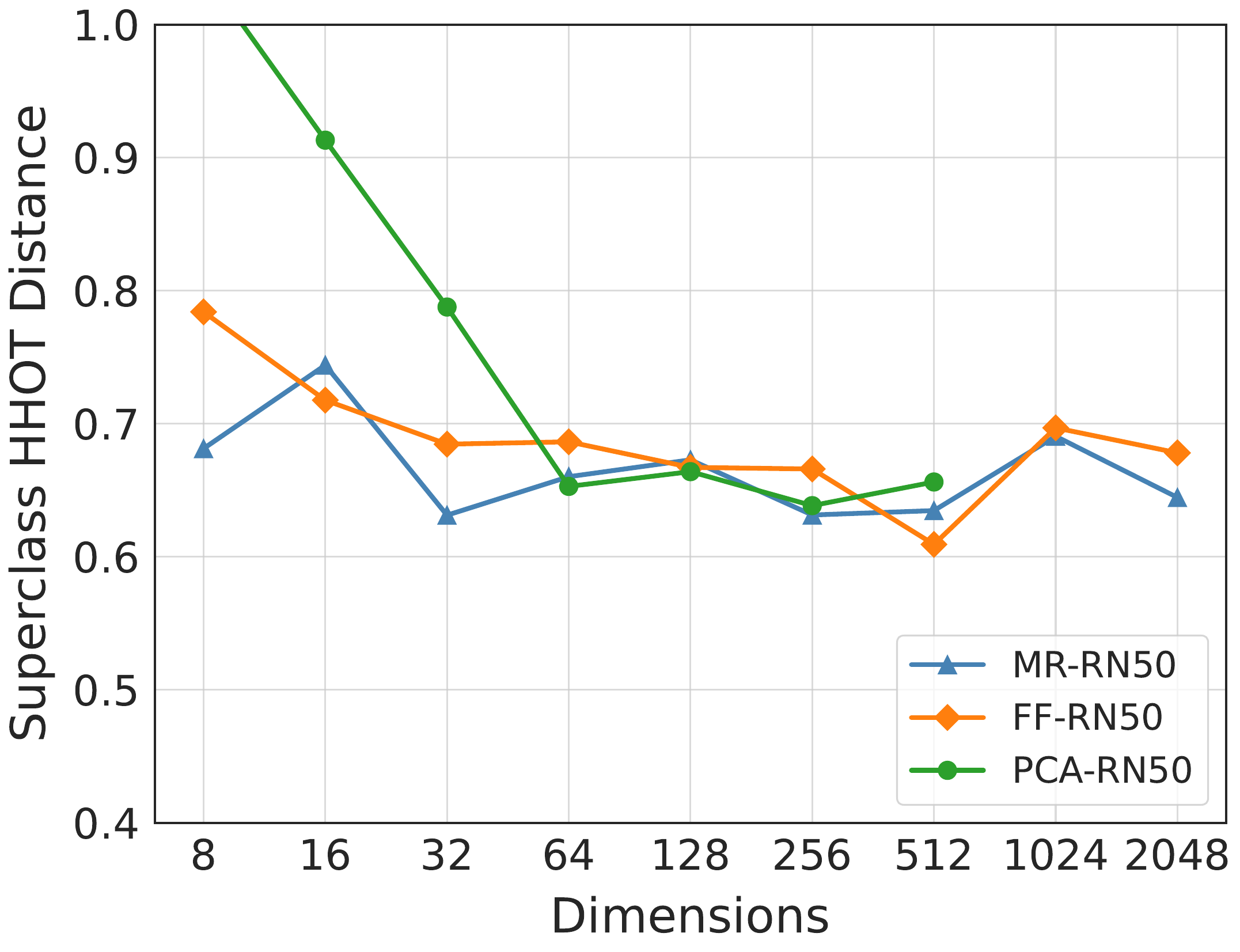}\qquad
          \includegraphics[width=0.3\linewidth]{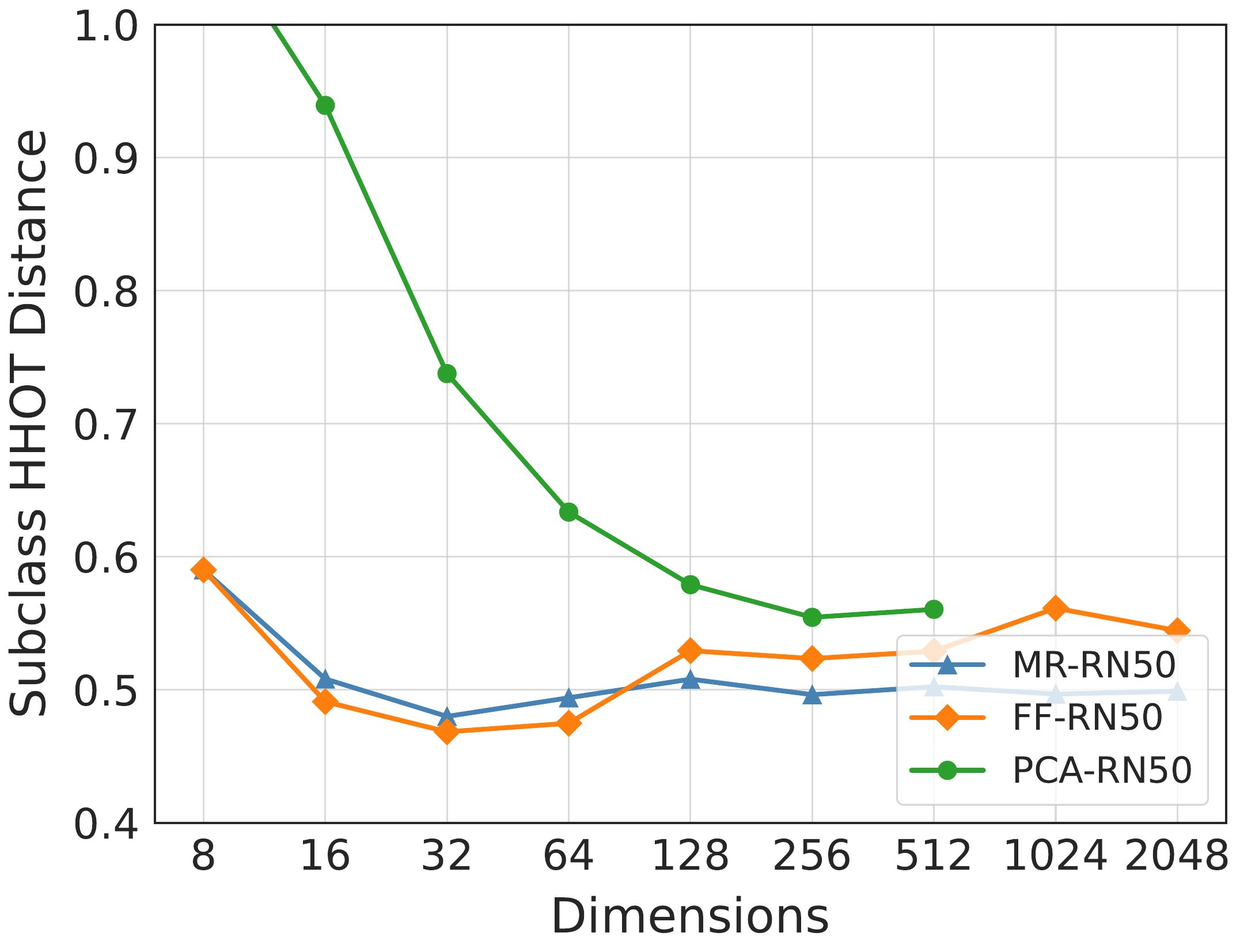}
      }\vspace{-2mm}
            \caption{Average HHOT distance between predicted superclass and subclass clusters and the true hierarchy shows that MR clusters align the best with the ground truth.}
                  \label{fig:c3}
                \vspace{-6mm}
    \end{figure}

On control datasets, MRs are more accurate than standard FF embeddings at all dimensions for both superclass and subclass classification, reflecting results from \citet{kusupati2022matryoshka} (Appendix \ref{apdx:b1}). In addition, MR has the highest purity and AMI for subclass clustering past 64 dimensions. However, there is no benefit for superclass clustering, where MR, FF, and PCA all perform equally (Figure \ref{fig:c2}). This suggests that in the control datasets, while MR is better at separating the classes the models were trained on, it is no better at identifying implicit hierarchical relationships between classes. Similarly, in Figure~\ref{fig:c3}, while the HHOT distance for MR's subclass clustering is the lowest of all dimensions, the gap between it and the other embeddings disappears for superclass clusters.

We also conducted the same experiments for the 4 fine-grained datasets and 5 high-variance datasets of HierNet. Embeddings on high-variance datasets exhibit a similar trend to that of the control datasets (Appendix \ref{apdx:b3}). However, on fine-grained datasets, MR outperforms FF and PCA on AMI, purity, and HHOT distance for both superclass and subclass clusterings. 
While MR embeddings do not show any general hierarchical advantage, MR embeddings seem to possess some benefit when classes become difficult to separate. Finally, we also note that lower-dimensional FF representations can also capture some hierarchy while PCA-style projections from high-dimensional representations might not preserve the underlying notion of hierarchy.
\section{Conclusions}
In this empirical investigation, we present a novel suite of hierarchical datasets and show that ``hierarchical'' embeddings are not inherently better at capturing underlying hierarchies in visual data. We also demonstrate that standard Euclidean embeddings are able to competitively capture hierarchy without explicit training to do so. However, hierarchical embeddings still stand to have an impact on other axes like interpretability through entailment in MERU and a significant reduction in search costs through Matryoshka Representations. In the future, we see the potential for our work to be used to benchmark and assist the development of new hierarchical representations that possess significant performance boosts over standard Euclidean/spherical representations.

\section{Acknowledgements}
This work is in part supported by NSF IIS 1652052, IIS 1703166, DARPA N66001-19-2-4031, DARPA W911NF-15-1-0543 and gifts from Allen Institute for Artificial Intelligence, Google and Apple. In addition, thanks to Aniket Rege for his help on this project.

\bibliography{pmlr-sample}

\begin{thebibliography}{12}
\providecommand{\natexlab}[1]{#1}
\providecommand{\url}[1]{\texttt{#1}}
\expandafter\ifx\csname urlstyle\endcsname\relax
  \providecommand{\doi}[1]{doi: #1}\else
  \providecommand{\doi}{doi: \begingroup \urlstyle{rm}\Url}\fi

\bibitem[Barz and Denzler(2019)]{barz2019hierarchy}
Bj{\"o}rn Barz and Joachim Denzler.
\newblock Hierarchy-based image embeddings for semantic image retrieval.
\newblock In \emph{2019 IEEE winter conference on applications of computer
  vision (WACV)}, pages 638--647. IEEE, 2019.

\bibitem[Desai et~al.(2021)Desai, Kaul, Aysola, and Johnson]{desai2021redcaps}
Karan Desai, Gaurav Kaul, Zubin Aysola, and Justin Johnson.
\newblock {RedCaps: Web-curated image-text data created by the people, for the
  people}.
\newblock In \emph{NeurIPS Datasets and Benchmarks}, 2021.

\bibitem[Desai et~al.(2023)Desai, Nickel, Rajpurohit, Johnson, and
  Vedantam]{desai2023meru}
Karan Desai, Maximilian Nickel, Tanmay Rajpurohit, Justin Johnson, and
  Ramakrishna Vedantam.
\newblock {Hyperbolic Image-Text Representations}.
\newblock In \emph{Proceedings of the International Conference on Machine
  Learning}, 2023.

\bibitem[Kusupati et~al.(2022)Kusupati, Bhatt, Rege, Wallingford, Sinha,
  Ramanujan, Howard-Snyder, Chen, Kakade, Jain, et~al.]{kusupati2022matryoshka}
Aditya Kusupati, Gantavya Bhatt, Aniket Rege, Matthew Wallingford, Aditya
  Sinha, Vivek Ramanujan, William Howard-Snyder, Kaifeng Chen, Sham Kakade,
  Prateek Jain, et~al.
\newblock Matryoshka representation learning.
\newblock In \emph{Advances in Neural Information Processing Systems}, December
  2022.

\bibitem[Miller(1995)]{miller1995wordnet}
George~A Miller.
\newblock Wordnet: a lexical database for english.
\newblock \emph{Communications of the ACM}, 38\penalty0 (11):\penalty0 39--41,
  1995.

\bibitem[Monath et~al.(2021)Monath, Dubey, Guruganesh, Zaheer, Ahmed, McCallum,
  Mergen, Najork, Terzihan, Tjanaka, et~al.]{monath2021scalable}
Nicholas Monath, Kumar~Avinava Dubey, Guru Guruganesh, Manzil Zaheer, Amr
  Ahmed, Andrew McCallum, Gokhan Mergen, Marc Najork, Mert Terzihan, Bryon
  Tjanaka, et~al.
\newblock Scalable hierarchical agglomerative clustering.
\newblock In \emph{Proceedings of the 27th ACM SIGKDD Conference on knowledge
  discovery \& data mining}, pages 1245--1255, 2021.

\bibitem[Nguyen and Kornblith(2023)]{nguyen2023probing}
Thao Nguyen and Simon Kornblith.
\newblock {Probing Clustering in Neural Network Representations}.
\newblock 2023.

\bibitem[Nickel and Kiela(2017)]{nickel2017poincare}
Maximillian Nickel and Douwe Kiela.
\newblock Poincar{\'e} embeddings for learning hierarchical representations.
\newblock \emph{Advances in neural information processing systems}, 30, 2017.

\bibitem[Radford et~al.(2021)Radford, Kim, Hallacy, Ramesh, Goh, Agarwal,
  Sastry, Askell, Mishkin, Clark, et~al.]{radford2021learning}
Alec Radford, Jong~Wook Kim, Chris Hallacy, Aditya Ramesh, Gabriel Goh,
  Sandhini Agarwal, Girish Sastry, Amanda Askell, Pamela Mishkin, Jack Clark,
  et~al.
\newblock Learning transferable visual models from natural language
  supervision.
\newblock In \emph{International conference on machine learning}, pages
  8748--8763. PMLR, 2021.

\bibitem[Russakovsky et~al.(2015)Russakovsky, Deng, Su, Krause, Satheesh, Ma,
  Huang, Karpathy, Khosla, Bernstein, et~al.]{russakovsky2015imagenet}
Olga Russakovsky, Jia Deng, Hao Su, Jonathan Krause, Sanjeev Satheesh, Sean Ma,
  Zhiheng Huang, Andrej Karpathy, Aditya Khosla, Michael Bernstein, et~al.
\newblock Imagenet large scale visual recognition challenge.
\newblock \emph{International journal of computer vision}, 115:\penalty0
  211--252, 2015.

\bibitem[Santurkar et~al.(2020)Santurkar, Tsipras, and
  Madry]{santurkar2020breeds}
Shibani Santurkar, Dimitris Tsipras, and Aleksander Madry.
\newblock Breeds: Benchmarks for subpopulation shift.
\newblock In \emph{ArXiv preprint arXiv:2008.04859}, 2020.

\bibitem[Yeaton et~al.(2022)Yeaton, Krishnan, Mieloszyk, Alvarez-Melis, and
  Huynh]{yeaton2022hierarchical}
Anna Yeaton, Rahul~G Krishnan, Rebecca Mieloszyk, David Alvarez-Melis, and
  Grace Huynh.
\newblock Hierarchical optimal transport for comparing histopathology datasets.
\newblock In \emph{Medical Imaging with Deep Learning}, 2022.
\newblock URL \url{https://openreview.net/forum?id=gADV7sV4CMo}.

\end{thebibliography}

\newpage
\appendix
\section{Metrics}
    \label{apdx:metrics}
    \subsection{Mutual Information}
    Mutual information is given by 
    $$MI(A, B) = \sum_{i = 1}^{|A|} \sum_{j = 1}^{|B|} \frac{|A_i \cap B_j|}{N} \log{\frac{N|A_i \cap B_j|}{|A_i||B_j|}}$$
    where $A$ and $B$ are the clusterings being compared, $N$ is the total number of elements, and $A_i$ is the $i$th cluster in clustering $A$. 
    \subsection{Purity}
    Purity is given by
    $$\text{purity} = \frac{1}{N}\sum_{i=1}^{|C|} \max_{j}{|C_i \cap G_j|}$$
    where $C$ is the discovered clustering, $C_i$ is the ith cluster in $C$, and $G_j$ is the $j$th ground truth class.
\newpage
\section{HierNet Datasets}
\label{apdx:dsets}
    \textbf{Control Datasets}
        \begin{table}[h!]
    \centering
    \caption{ds5\_r2\_l5}
    \resizebox{0.8\textwidth}{!}{
        \begin{tabular}{lll}
        \toprule
        {} &                                    superclass &                                        subclasses (class number) \\
        \midrule
        0  &                    audio system, sound system &                   [loudspeaker (632), iPod (605)] \\
        1  &                                          bowl &              [mixing bowl (659), soup bowl (809)] \\
        2  &                   camera, photographic camera &      [Polaroid camera (732), reflex camera (759)] \\
        3  &                              digital computer &        [hand-held computer (590), notebook (681)] \\
        4  &                     firearm, piece, small-arm &                     [rifle (764), revolver (763)] \\
        5  &                         glass, drinking glass &                  [beer glass (441), goblet (572)] \\
        6  &                gymnastic apparatus, exerciser &        [horizontal bar (602), balance beam (416)] \\
        7  &                                           jug &              [whiskey jug (901), water jug (899)] \\
        8  &                                          lock &           [combination lock (507), padlock (695)] \\
        9  &                                           pen &                    [quill (749), ballpoint (418)] \\
        10 &  percussion instrument, percussive instrument &                   [steel drum (822), chime (494)] \\
        11 &                                           pot &                   [coffeepot (505), teapot (849)] \\
        12 &                           stringed instrument &    [acoustic guitar (402), electric guitar (546)] \\
        13 &               telephone, phone, telephone set &  [dial telephone (528), cellular telephone (487)] \\
        14 &               timepiece, timekeeper, horologe &                  [sundial (835), hourglass (604)] \\
        15 &        weight, free weight, exercising weight &                   [barbell (422), dumbbell (543)] \\
        16 &                         wind instrument, wind &                [French horn (566), bassoon (432)] \\
        \bottomrule
        \end{tabular}
    }
    
\end{table}

\begin{table}[h!]
    \centering
        \caption{ds6\_r2\_l5}
    \resizebox{0.8\textwidth}{!}{
    \begin{tabular}{lll}
    \toprule
    {} &                                         superclass &                                   subclasses \\
    \midrule
    0  &                                                bag &             [plastic bag (728), purse (748)] \\
    1  &  body armor, body armour, suit of armor, suit o... &  [breastplate (461), bulletproof vest (465)] \\
    2  &                                                cap &        [shower cap (793), mortarboard (667)] \\
    3  &                                               coat &            [trench coat (869), kimono (614)] \\
    4  &                         cream, ointment, emollient &              [lotion (631), sunscreen (838)] \\
    5  &                                          face mask &                  [mask (643), gasmask (570)] \\
    6  &                                  hat, chapeau, lid &               [bonnet (452), bearskin (439)] \\
    7  &                                             helmet &      [pickelhaube (715), crash helmet (518)] \\
    8  &                         makeup, make-up, war paint &          [lipstick (629), face powder (551)] \\
    9  &                                       necktie, tie &          [Windsor tie (906), bolo tie (451)] \\
    10 &                                              scarf &             [stole (824), feather boa (552)] \\
    11 &                                             sheath &              [scabbard (777), holster (597)] \\
    12 &                                               shoe &           [Loafer (630), running shoe (770)] \\
    13 &                                              skirt &           [hoopskirt (601), overskirt (689)] \\
    14 &                                    sweater, jumper &           [sweatshirt (841), cardigan (474)] \\
    15 &  swimsuit, swimwear, bathing suit, swimming cos... &       [maillot (639), swimming trunks (842)] \\
    16 &                        undergarment, unmentionable &              [diaper (529), brassiere (459)] \\
    \bottomrule
    \end{tabular}}

\end{table}
    
\begin{table}[h!]
    \centering
        \caption{ds7\_r2\_l5}
    \resizebox{0.8\textwidth}{!}{
    \begin{tabular}{llll}
    \toprule
    {} &                                         superclass &                                subclasses \\
    \midrule
    0  &                                         salamander &                           [eft (27), axolotl (29)] \\
    1  &                                             turtle &         [box turtle (37), leatherback turtle (34)] \\
    2  &                                             lizard &             [whiptail (41), alligator lizard (44)] \\
    3  &                           snake, serpent, ophidian &              [night snake (60), garter snake (57)] \\
    4  &                                             spider &  [tarantula (76), black and gold garden spider ...  \\
    5  &                                             grouse &             [ptarmigan (81), prairie chicken (83)] \\
    6  &                                             parrot &                        [macaw (88), lorikeet (90)] \\
    7  &                                               crab &         [Dungeness crab (118), fiddler crab (120)] \\
    8  &                dog, domestic dog, Canis familiaris &                 [bloodhound (163), Pekinese (154)] \\
    9  &                                               wolf &                     [coyote (272), red wolf (271)] \\
    10 &                                                fox &                 [grey fox (280), Arctic fox (279)] \\
    11 &  domestic cat, house cat, Felis domesticus, Fel... &              [tiger cat (282), Egyptian cat (285)] \\
    12 &                                               bear &      [sloth bear (297), American black bear (295)] \\
    13 &                                             beetle &       [dung beetle (305), rhinoceros beetle (306)] \\
    14 &                                          butterfly &           [sulphur butterfly (325), admiral (321)] \\
    15 &                                                ape &                    [gibbon (368), orangutan (365)] \\
    16 &                                             monkey &                       [marmoset (377), titi (380)]\\
    \bottomrule
    \end{tabular}}

\end{table}
        \newpage
    \textbf{Fine-Grained Datasets}
        \begin{table}[h!]
    \centering
        \caption{ds8\_r3\_l5}
    \resizebox{0.8\textwidth}{!}{
    \begin{tabular}{lll}
    \toprule
    {} &                                         superclass &                                       subclasses \\
    \midrule
    0 &                                              whale &           [killer whale (148), grey whale (147)] \\
    1 &                dog, domestic dog, Canis familiaris &                [cairn (192), Newfoundland (256)] \\
    2 &                                               wolf &              [timber wolf (269), red wolf (271)] \\
    3 &                                           wild dog &                       [dingo (273), dhole (274)] \\
    4 &  domestic cat, house cat, Felis domesticus, Fel... &          [Egyptian cat (285), Persian cat (283)] \\
    5 &                                               bear &               [brown bear (294), ice bear (296)] \\
    6 &                                rabbit, coney, cony &                [Angora (332), wood rabbit (330)] \\
    7 &                                                ape &                [chimpanzee (367), gorilla (366)] \\
    8 &                                             monkey &            [squirrel monkey (382), guenon (370)] \\
    9 &                                           elephant &  [African elephant (386), Indian elephant (385)] \\
    \bottomrule
    \end{tabular}}

\end{table}

\begin{table}[h!]
    \centering
        \caption{ds9\_r3\_l5}
    \resizebox{0.8\textwidth}{!}{
    \begin{tabular}{lll}
    \toprule
    {} &                                         superclass &                                   subclasses \\
    \midrule
    0 &                                                bag &             [plastic bag (728), purse (748)] \\
    1 &  body armor, body armour, suit of armor, suit o... &  [breastplate (461), bulletproof vest (465)] \\
    2 &                                                cap &        [shower cap (793), mortarboard (667)] \\
    3 &                                          face mask &                  [gasmask (570), mask (643)] \\
    4 &                                  hat, chapeau, lid &           [bearskin (439), cowboy hat (515)] \\
    5 &                                             helmet &  [football helmet (560), crash helmet (518)] \\
    6 &                                       necktie, tie &           [bow tie (457), Windsor tie (906)] \\
    7 &                                              scarf &             [feather boa (552), stole (824)] \\
    8 &                                             sheath &              [holster (597), scabbard (777)] \\
    9 &                                               shoe &           [running shoe (770), Loafer (630)] \\
    \bottomrule
    \end{tabular}}

\end{table}

\begin{table}[h!]
    \centering
        \caption{ds10\_r3\_l5}
    \resizebox{0.8\textwidth}{!}{
    \begin{tabular}{lll}
    \toprule
    {} &                       superclass &                               subclasses \\
    \midrule
    0 &                             bowl &     [soup bowl (809), mixing bowl (659)] \\
    1 &            glass, drinking glass &         [goblet (572), beer glass (441)] \\
    2 &                              jug &     [water jug (899), whiskey jug (901)] \\
    3 &              keyboard instrument &         [accordion (401), upright (881)] \\
    4 &                             lock &  [combination lock (507), padlock (695)] \\
    5 &                           opener &      [can opener (473), corkscrew (512)] \\
    6 &                 pan, cooking pan &            [frying pan (567), wok (909)] \\
    7 &                              pot &         [coffeepot (505), caldron (469)] \\
    8 &              stringed instrument &      [harp (594), electric guitar (546)] \\
    9 &  timepiece, timekeeper, horologe &   [wall clock (892), analog clock (409)] \\
    \bottomrule
    \end{tabular}}
\end{table}
    
\begin{table}[h!]
    \centering
        \caption{ds11\_r3\_l5}
    \resizebox{0.8\textwidth}{!}{
    \begin{tabular}{lll}
    \toprule
    {} &                                         superclass &                                 subclasses \\
    \midrule
    0 &                                       bridge, span &   [viaduct (888), suspension bridge (839)] \\
    1 &                                     column, pillar &          [totem pole (863), obelisk (682)] \\
    2 &  dwelling, home, domicile, abode, habitation, d... &            [monastery (663), castle (483)] \\
    3 &                                     fence, fencing &  [chainlink fence (489), worm fence (912)] \\
    4 &                                 memorial, monument &        [brass (458), triumphal arch (873)] \\
    5 &  mercantile establishment, retail store, sales ... &        [barbershop (424), shoe shop (788)] \\
    6 &                                        outbuilding &                 [barn (425), apiary (410)] \\
    7 &  place of worship, house of prayer, house of Go... &                [stupa (832), church (497)] \\
    8 &                                               roof &             [tile roof (858), vault (884)] \\
    9 &                                    signboard, sign &   [street sign (919), traffic light (920)] \\
    \bottomrule
    \end{tabular}}

\end{table}
        \newpage
    \textbf{High-Variance Datasets}
        \begin{table}[h!]
    \centering
     \caption{ds0\_r0\_l5}
    \resizebox{0.8\textwidth}{!}{
    \begin{tabular}{lll}
    \toprule
    {} &                              superclass &                                       subclasses \\
    \midrule
    0  &                snake, serpent, ophidian &             [vine snake (59), Indian cobra (63)] \\
    1  &                                 lobster &    [spiny lobster (123), American lobster (122)] \\
    2  &                               sandpiper &     [red-backed sandpiper (140), redshank (141)] \\
    3  &                                     ape &                 [chimpanzee (367), gibbon (368)] \\
    4  &                                  monkey &                     [baboon (372), guenon (370)] \\
    5  &                                   lemur &              [indri (384), Madagascar cat (383)] \\
    6  &                                elephant &  [Indian elephant (385), African elephant (386)] \\
    7  &   curtain, drape, drapery, mantle, pall &    [shower curtain (794), theater curtain (854)] \\
    8  &               firearm, piece, small-arm &               [assault rifle (413), rifle (764)] \\
    9  &       handcart, pushcart, cart, go-cart &              [barrow (428), shopping cart (791)] \\
    10 &              makeup, make-up, war paint &              [lipstick (629), face powder (551)] \\
    11 &                     sofa, couch, lounge &           [studio couch (831), park bench (703)] \\
    12 &                                   towel &            [bath towel (434), paper towel (700)] \\
    13 &                       truck, motortruck &            [moving van (675), fire engine (555)] \\
    14 &  weight, free weight, exercising weight &                  [barbell (422), dumbbell (543)] \\
    15 &                   wind instrument, wind &               [panpipe (699), French horn (566)] \\
    16 &                                   sauce &         [chocolate sauce (960), carbonara (959)] \\
    \bottomrule
    \end{tabular}}
   
\end{table}

\begin{table}[h!]
    \centering
    \caption{ds1\_r0\_l5}
     \resizebox{0.8\textwidth}{!}{
    \begin{tabular}{lll}
    \toprule
    {} &                                         superclass &                                         subclasses \\
    \midrule
    0  &                                             turtle &                 [box turtle (37), loggerhead (33)] \\
    1  &                                               crab &         [fiddler crab (120), Dungeness crab (118)] \\
    2  &                                              stork &             [black stork (128), white stork (127)] \\
    3  &                                          butterfly &           [sulphur butterfly (325), admiral (321)] \\
    4  &                        bicycle, bike, wheel, cycle &  [mountain bike (671), bicycle-built-for-two (4... \\
    5  &                                             bottle &            [water bottle (898), wine bottle (907)] \\
    6  &                         cream, ointment, emollient &                    [lotion (631), sunscreen (838)] \\
    7  &                          firearm, piece, small-arm &                 [assault rifle (413), rifle (764)] \\
    8  &                                                jug &               [water jug (899), whiskey jug (901)] \\
    9  &       percussion instrument, percussive instrument &                     [drum (541), steel drum (822)] \\
    10 &                                               ship &              [aircraft carrier (403), wreck (913)] \\
    11 &                                    signboard, sign &              [street sign (919), scoreboard (781)] \\
    12 &                                stringed instrument &                          [harp (594), cello (486)] \\
    13 &  swimsuit, swimwear, bathing suit, swimming cos... &                      [bikini (445), maillot (639)] \\
    14 &                    telephone, phone, telephone set &        [cellular telephone (487), pay-phone (707)] \\
    15 &                    timepiece, timekeeper, horologe &             [digital watch (531), stopwatch (826)] \\
    16 &                                             squash &           [spaghetti squash (940), zucchini (939)] \\
    \bottomrule
    \end{tabular}}
    
\end{table}

\begin{table}[h!]
    \centering
    \caption{ds2\_r0\_l5}
    \resizebox{0.8\textwidth}{!}{
    \begin{tabular}{lll}
    \toprule
    {} &                               superclass &                                     subclasses \\
    \midrule
    0  &                                phasianid &                     [peacock (84), quail (85)] \\
    1  &                                   parrot &             [African grey (87), lorikeet (90)] \\
    2  &                                     duck &      [red-breasted merganser (98), drake (97)] \\
    3  &                                butterfly &       [sulphur butterfly (325), admiral (321)] \\
    4  &                      rabbit, coney, cony &              [Angora (332), wood rabbit (330)] \\
    5  &                                 antelope &               [hartebeest (351), impala (352)] \\
    6  &                                      ape &               [chimpanzee (367), gibbon (368)] \\
    7  &                                   monkey &                   [baboon (372), guenon (370)] \\
    8  &                     baby bed, baby's bed &                     [crib (520), cradle (516)] \\
    9  &  car, railcar, railway car, railroad car &       [passenger car (705), freight car (565)] \\
    10 &    curtain, drape, drapery, mantle, pall &  [shower curtain (794), theater curtain (854)] \\
    11 &           gymnastic apparatus, exerciser &      [balance beam (416), parallel bars (702)] \\
    12 &                       memorial, monument &                  [brass (458), megalith (649)] \\
    13 &                                      pot &               [caldron (469), coffeepot (505)] \\
    14 &                      sofa, couch, lounge &         [studio couch (831), park bench (703)] \\
    15 &                        truck, motortruck &          [moving van (675), fire engine (555)] \\
    16 &                                   squash &       [spaghetti squash (940), zucchini (939)] \\
    \bottomrule
    \end{tabular}}
    
\end{table}

\begin{table}[h!]
    \centering
    \caption{ds3\_r0\_l5}
    \resizebox{0.8\textwidth}{!}{
    \begin{tabular}{lll}
    \toprule
    {} &                                         superclass &                                     subclasses \\
    \midrule
    0  &                                         salamander &               [axolotl (29), common newt (26)] \\
    1  &                           snake, serpent, ophidian &           [vine snake (59), Indian cobra (63)] \\
    2  &                                           wild dog &                     [dingo (273), dhole (274)] \\
    3  &  domestic cat, house cat, Felis domesticus, Fel... &           [tiger cat (282), Siamese cat (284)] \\
    4  &                                               bear &  [sloth bear (297), American black bear (295)] \\
    5  &                               baby bed, baby's bed &                     [crib (520), cradle (516)] \\
    6  &                        camera, photographic camera &   [reflex camera (759), Polaroid camera (732)] \\
    7  &                          firearm, piece, small-arm &             [assault rifle (413), rifle (764)] \\
    8  &                  handcart, pushcart, cart, go-cart &            [barrow (428), shopping cart (791)] \\
    9  &                                                jug &           [water jug (899), whiskey jug (901)] \\
    10 &                                             opener &            [corkscrew (512), can opener (473)] \\
    11 &                                    signboard, sign &          [street sign (919), scoreboard (781)] \\
    12 &                                sofa, couch, lounge &         [studio couch (831), park bench (703)] \\
    13 &                        undergarment, unmentionable &                [diaper (529), brassiere (459)] \\
    14 &             weight, free weight, exercising weight &                [barbell (422), dumbbell (543)] \\
    15 &                              wind instrument, wind &             [panpipe (699), French horn (566)] \\
    16 &                                     frozen dessert &             [ice cream (928), ice lolly (929)] \\
    \bottomrule
    \end{tabular}}
    
\end{table}
    
\begin{table}[h!]
    \centering
    \caption{ds4\_r0\_l5}
    \resizebox{0.8\textwidth}{!}{
    \begin{tabular}{lll}
    \toprule
    {} &                                         superclass &                                         subclasses \\
    \midrule
    0  &                                          phasianid &                         [peacock (84), quail (85)] \\
    1  &                                               duck &          [red-breasted merganser (98), drake (97)] \\
    2  &                                          sandpiper &       [red-backed sandpiper (140), redshank (141)] \\
    3  &                dog, domestic dog, Canis familiaris &  [Australian terrier (193), Norwegian elkhound ... \\
    4  &                                           wild dog &                         [dingo (273), dhole (274)] \\
    5  &                                             monkey &                       [baboon (372), guenon (370)] \\
    6  &                                           elephant &    [Indian elephant (385), African elephant (386)] \\
    7  &              curtain, drape, drapery, mantle, pall &      [shower curtain (794), theater curtain (854)] \\
    8  &  dwelling, home, domicile, abode, habitation, d... &               [castle (483), cliff dwelling (500)] \\
    9  &                              glass, drinking glass &                   [beer glass (441), goblet (572)] \\
    10 &                     gymnastic apparatus, exerciser &          [balance beam (416), parallel bars (702)] \\
    11 &                                  hat, chapeau, lid &                   [bonnet (452), cowboy hat (515)] \\
    12 &                                        outbuilding &                    [boathouse (449), apiary (410)] \\
    13 &                                                pen &                  [fountain pen (563), quill (749)] \\
    14 &  place of worship, house of prayer, house of Go... &                       [mosque (668), church (497)] \\
    15 &                                              towel &              [bath towel (434), paper towel (700)] \\
    16 &             weight, free weight, exercising weight &                    [barbell (422), dumbbell (543)] \\
    \bottomrule
    \end{tabular}}
    
\end{table}
        \FloatBarrier

\newpage

\section{Matryoshka Representations}
\subsection{Accuracies}
    \label{apdx:b1}
    \begin{figure}[h!]
        \centering
          
          {%
              \includegraphics[width=0.3\linewidth]{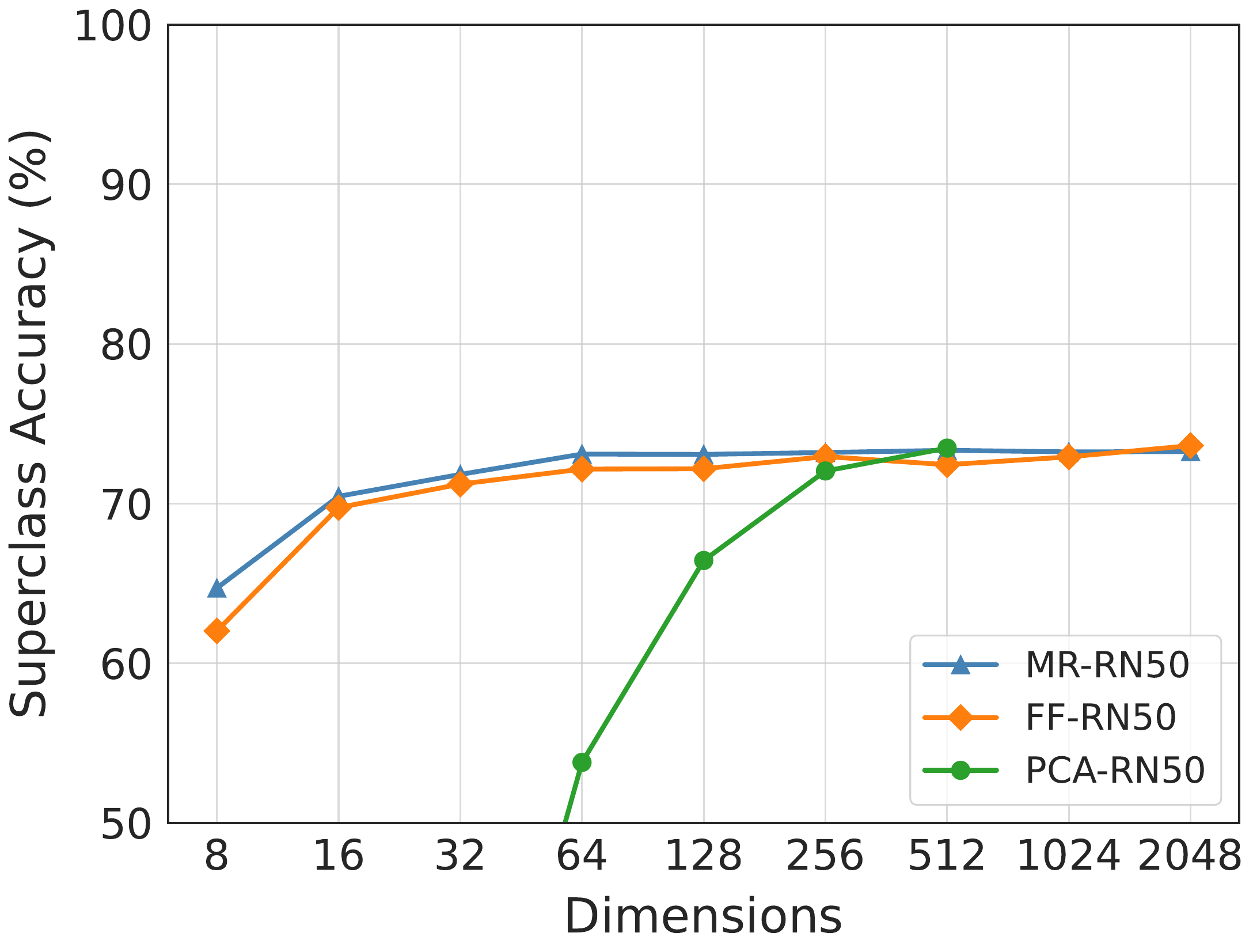}
            \qquad
              \includegraphics[width=0.3\linewidth]{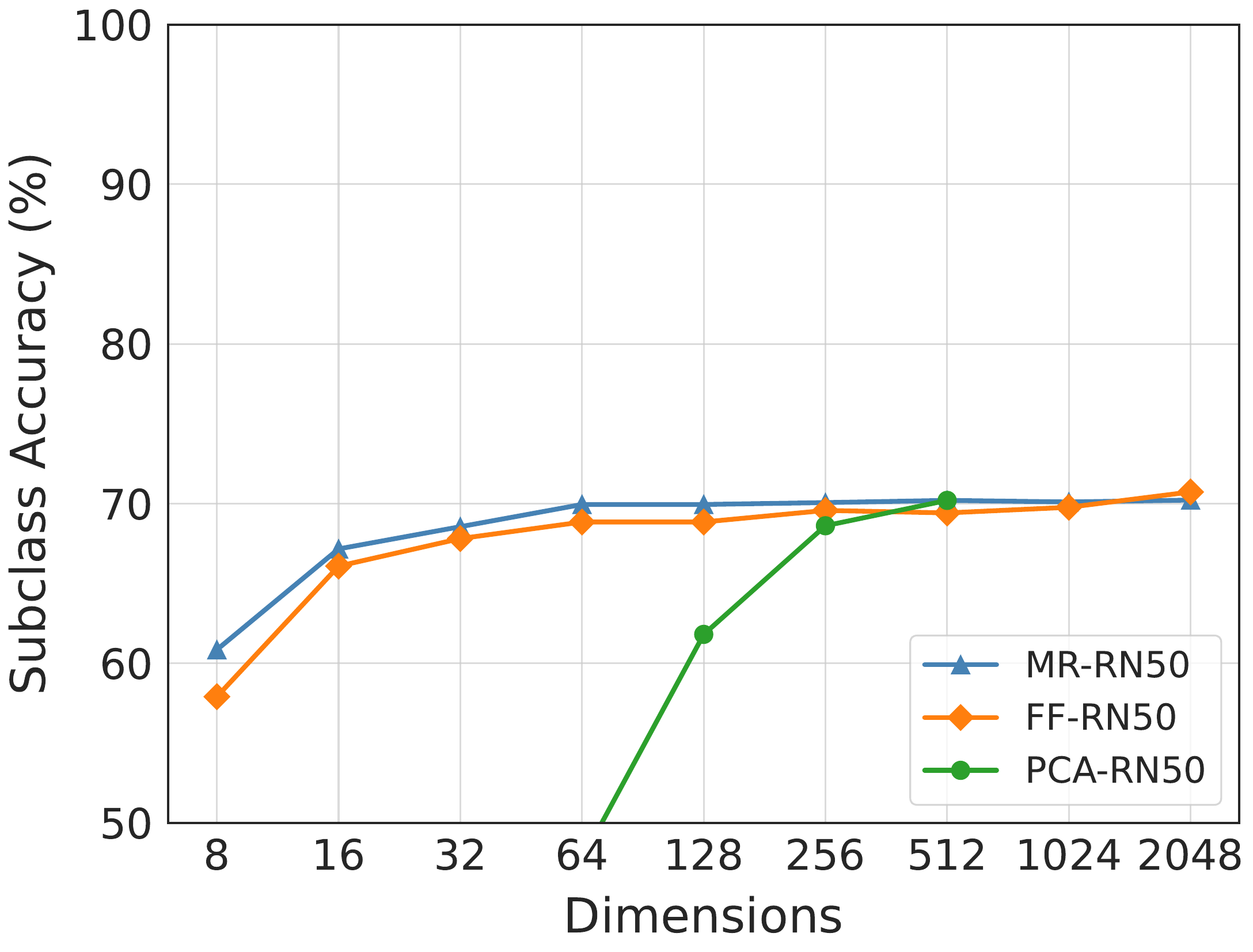}
          }
          \caption{Across control datasets, MR is slightly more accurate than FF when averaged across all three control datasets. Both embeddings are more accurate than PCA until 512 dimensions.}
          \label{fig:c1}
          \vspace{-6mm}
    \end{figure}
        \begin{figure}[htbp]
        \centering

          {%
              \includegraphics[width=0.4\linewidth]{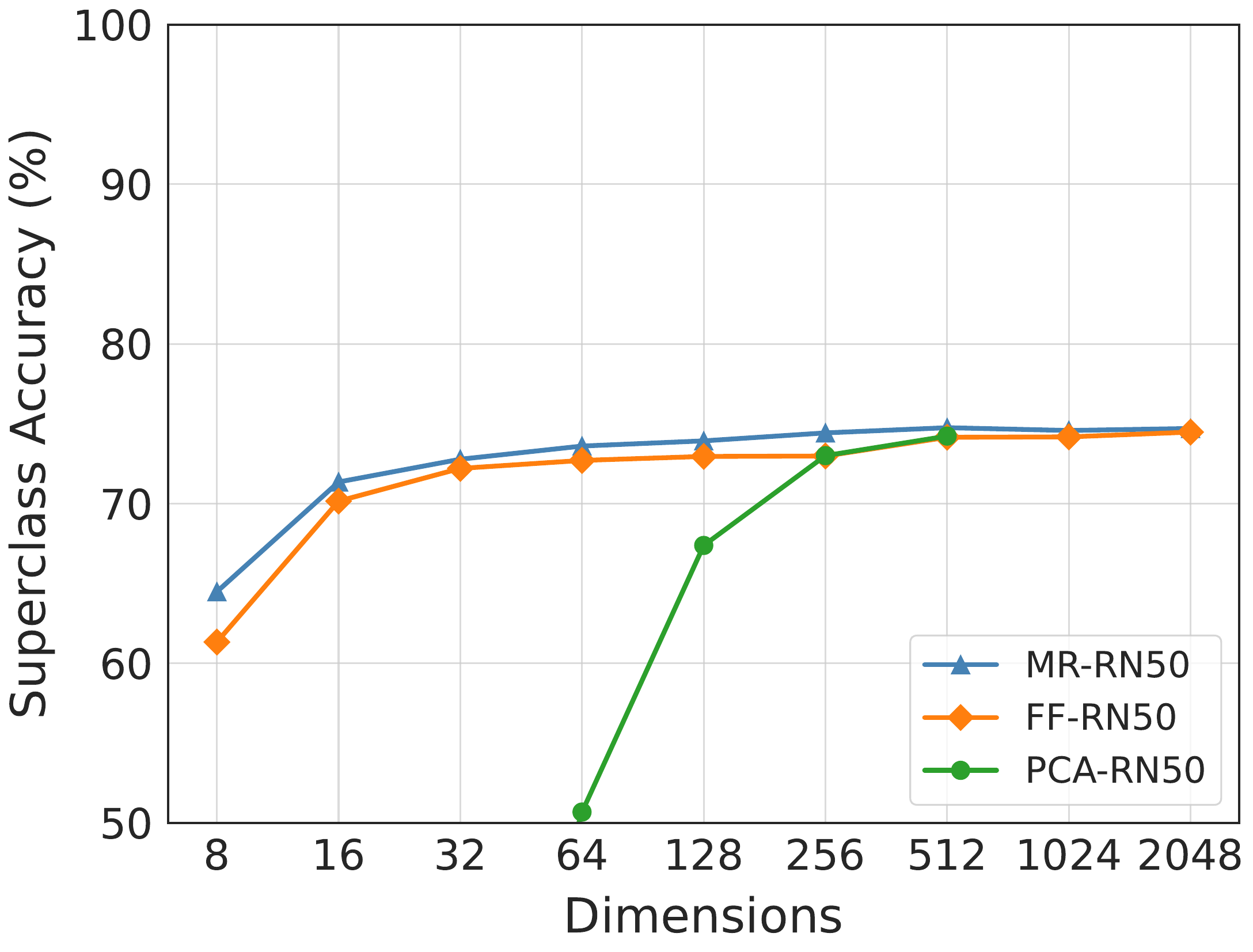}
            \qquad
              \includegraphics[width=0.4\linewidth]{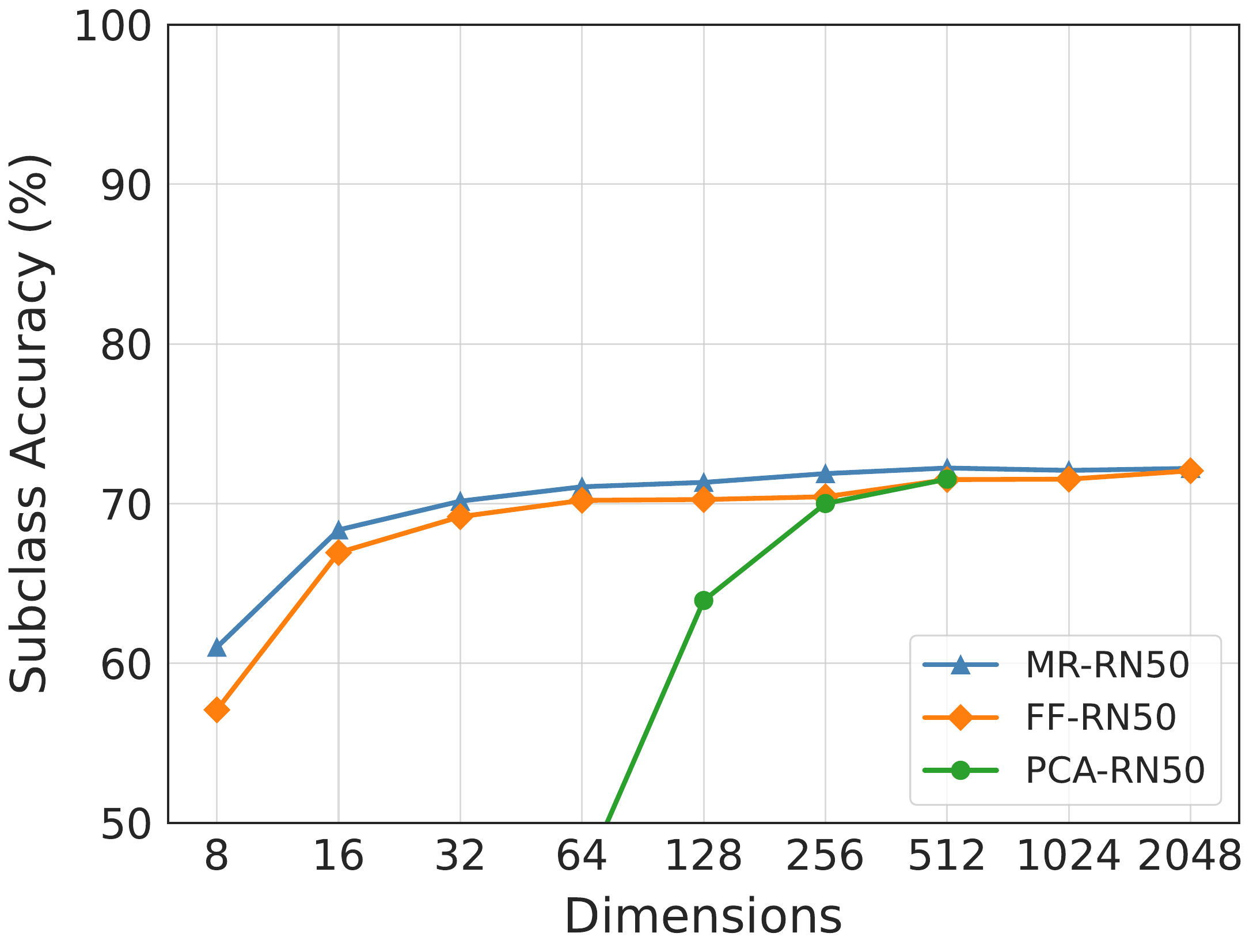}
          }
          \caption{In fine grained datasets, MR is slightly more accurate than FF.}
            \label{fig:f1}
    \end{figure}
    \begin{figure}[h!]
        \centering
          
          {%
              \includegraphics[width=0.4\linewidth]{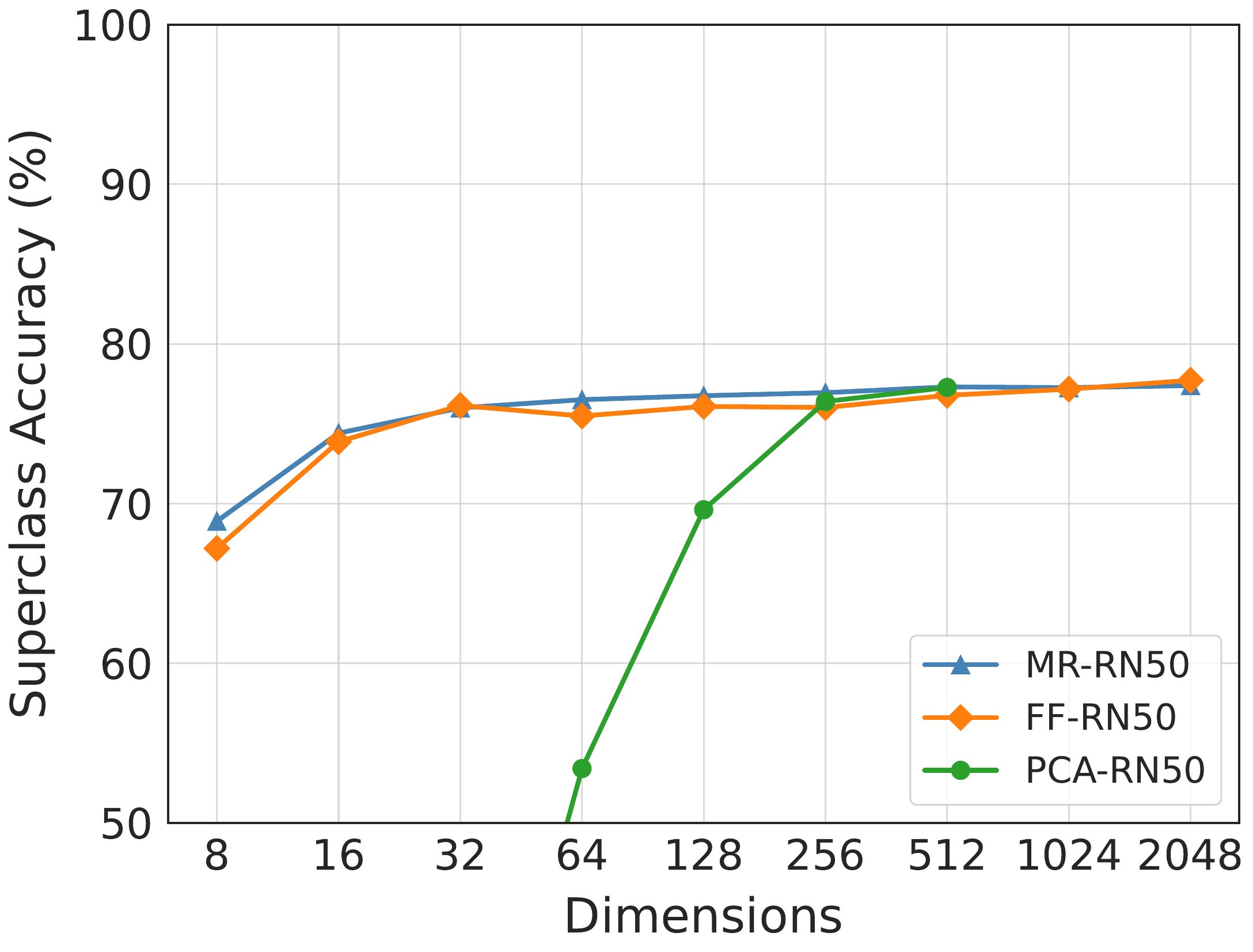}
            \qquad
              \includegraphics[width=0.4\linewidth]{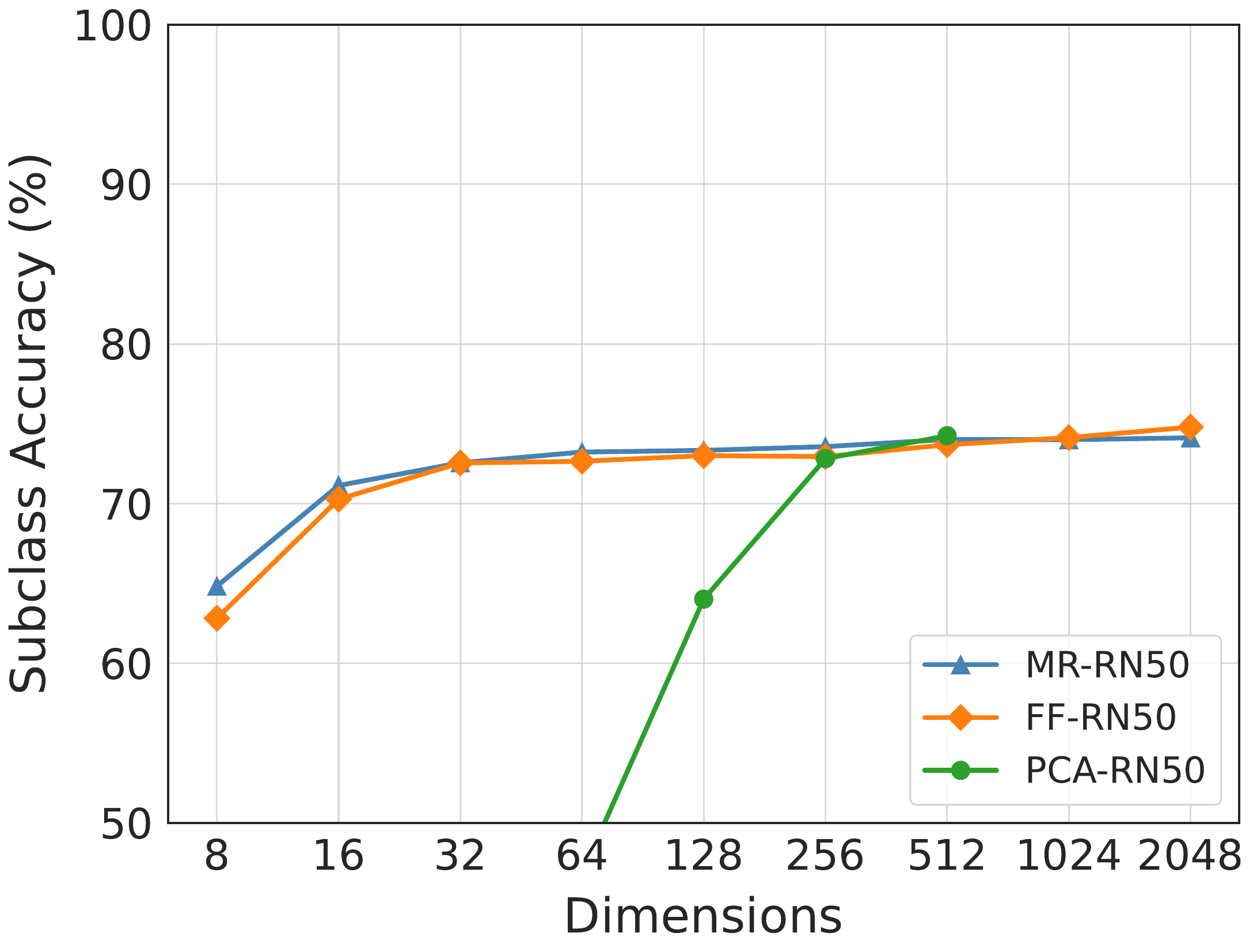}
          }
          \caption{In high-variance datasets, while MR is still the most accurate, the gap between it and other embeddings is smaller than before. We posit that this is due to high-variance datasets having very diverse and thus easily separable classes. This reduces the benefits of MR's information packing.}
          \label{fig:h1}
    \end{figure}
\newpage

\subsection{Fine-Grained Dataset Clustering}
    \label{apdx:b2}
    \begin{figure}[h!]
       \centering

      {%
          \includegraphics[width=0.22\linewidth]{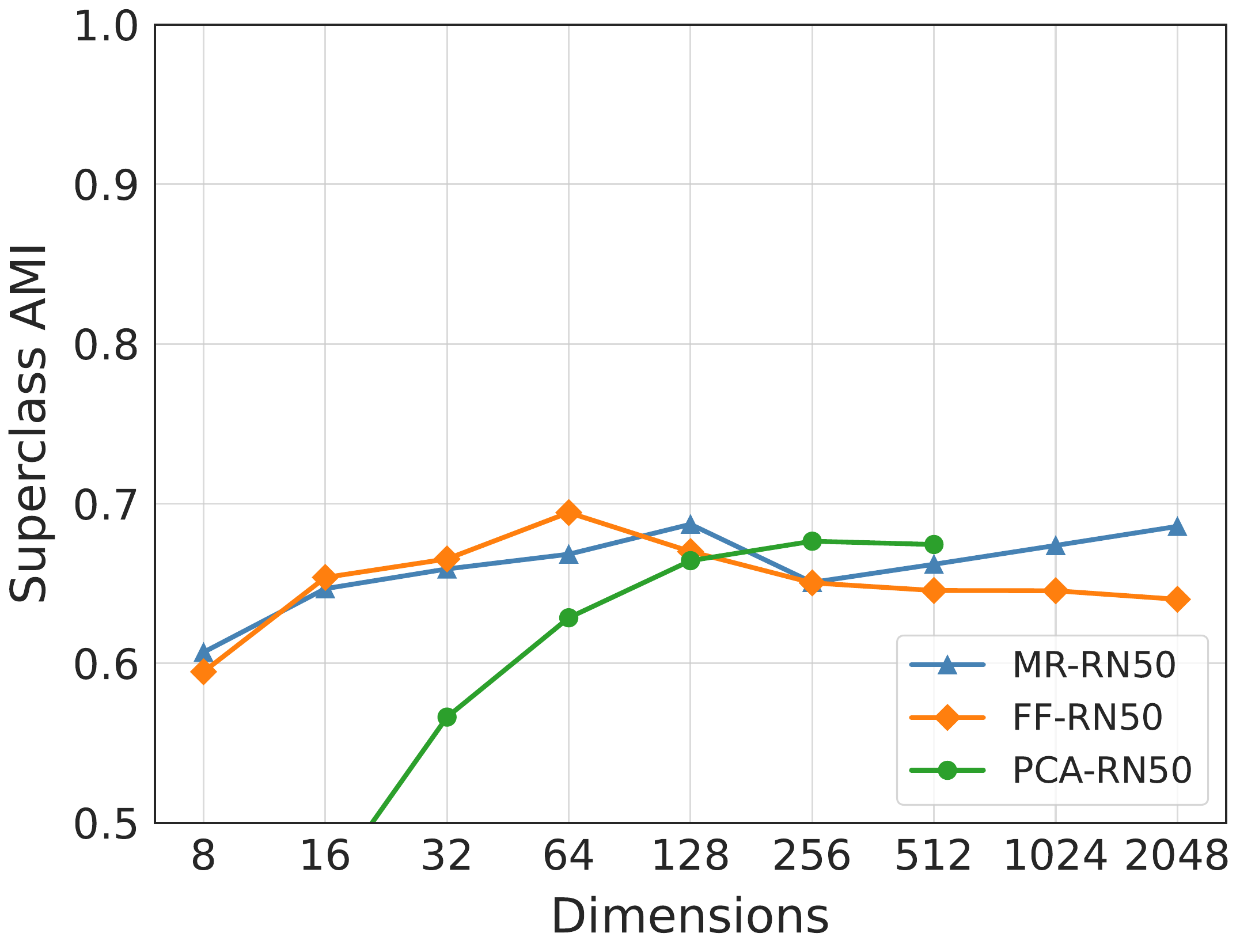}
          \includegraphics[width=0.22\linewidth]{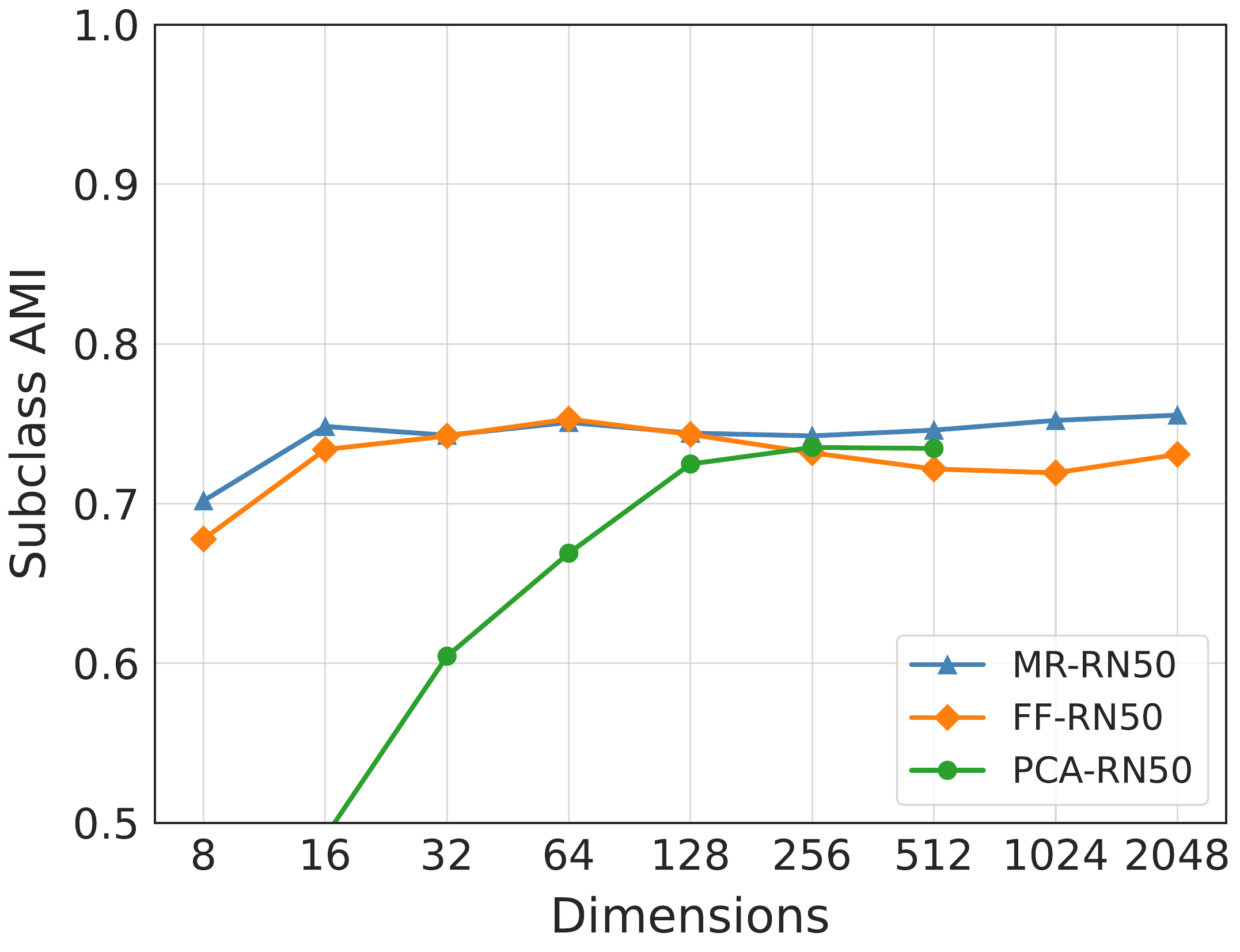}
          \includegraphics[width=0.22\linewidth]{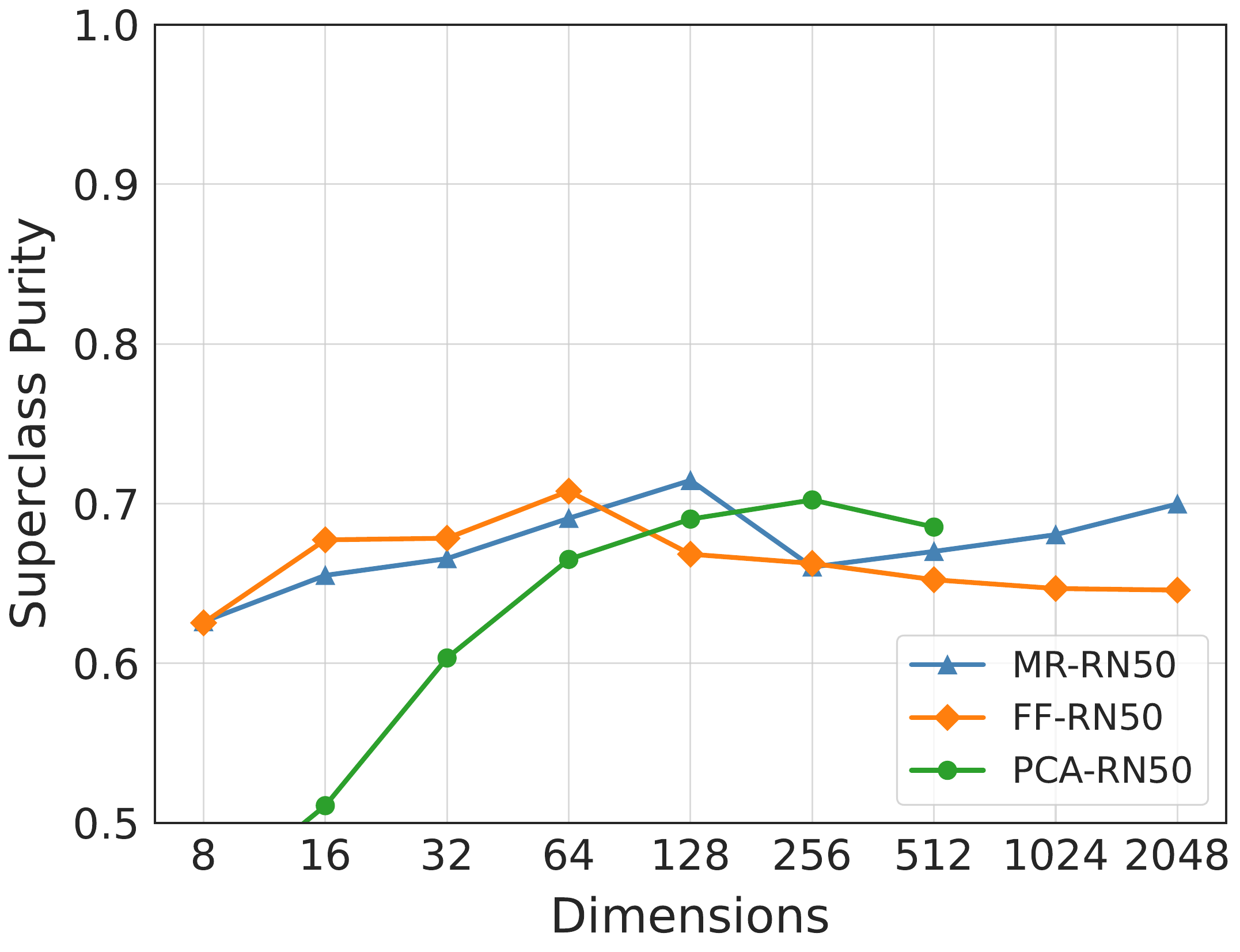}
          \includegraphics[width=0.22\linewidth]{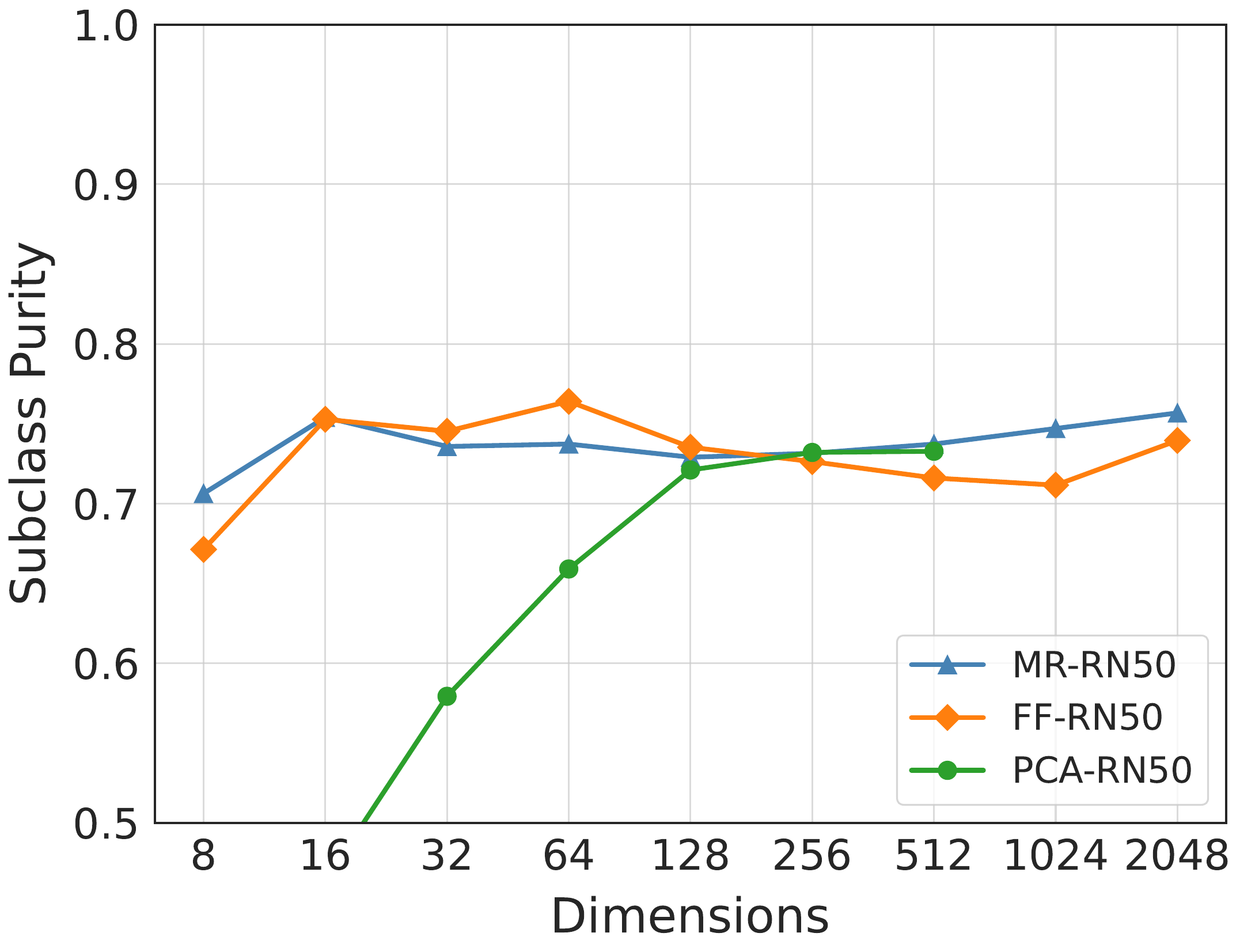}
      }
       \caption{In fine grained datasets, MR has consistently higher superclass and subclass AMI/purity than FF after 128 dimensions. However, MR's benefit over PCA is nonexistent between 128-d and 512-d.}
      \label{fig:f2}
    \end{figure}
    \begin{figure}[h!]
       
      \centering
      {%
          \includegraphics[width=0.4\linewidth]{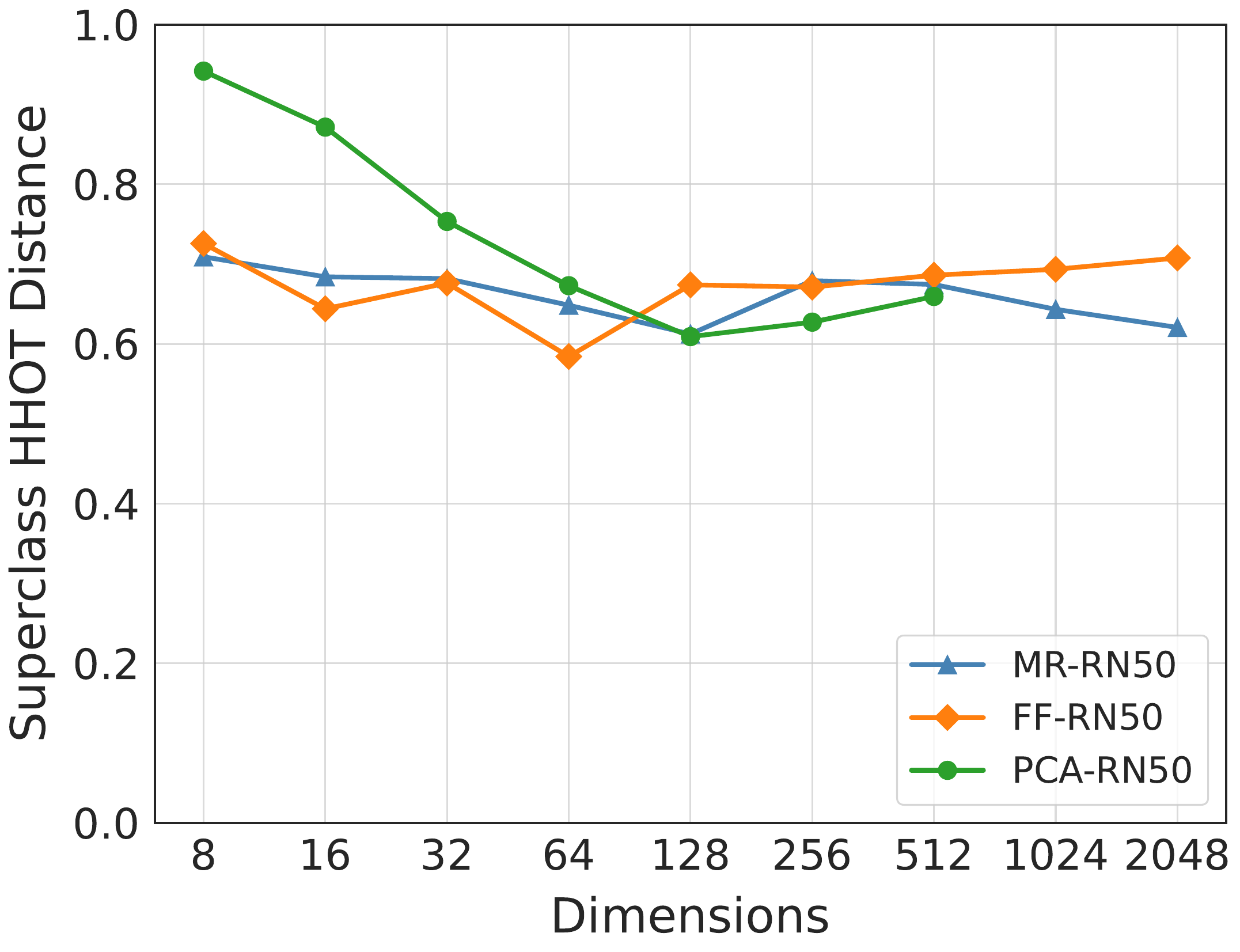}
          \includegraphics[width=0.4\linewidth]{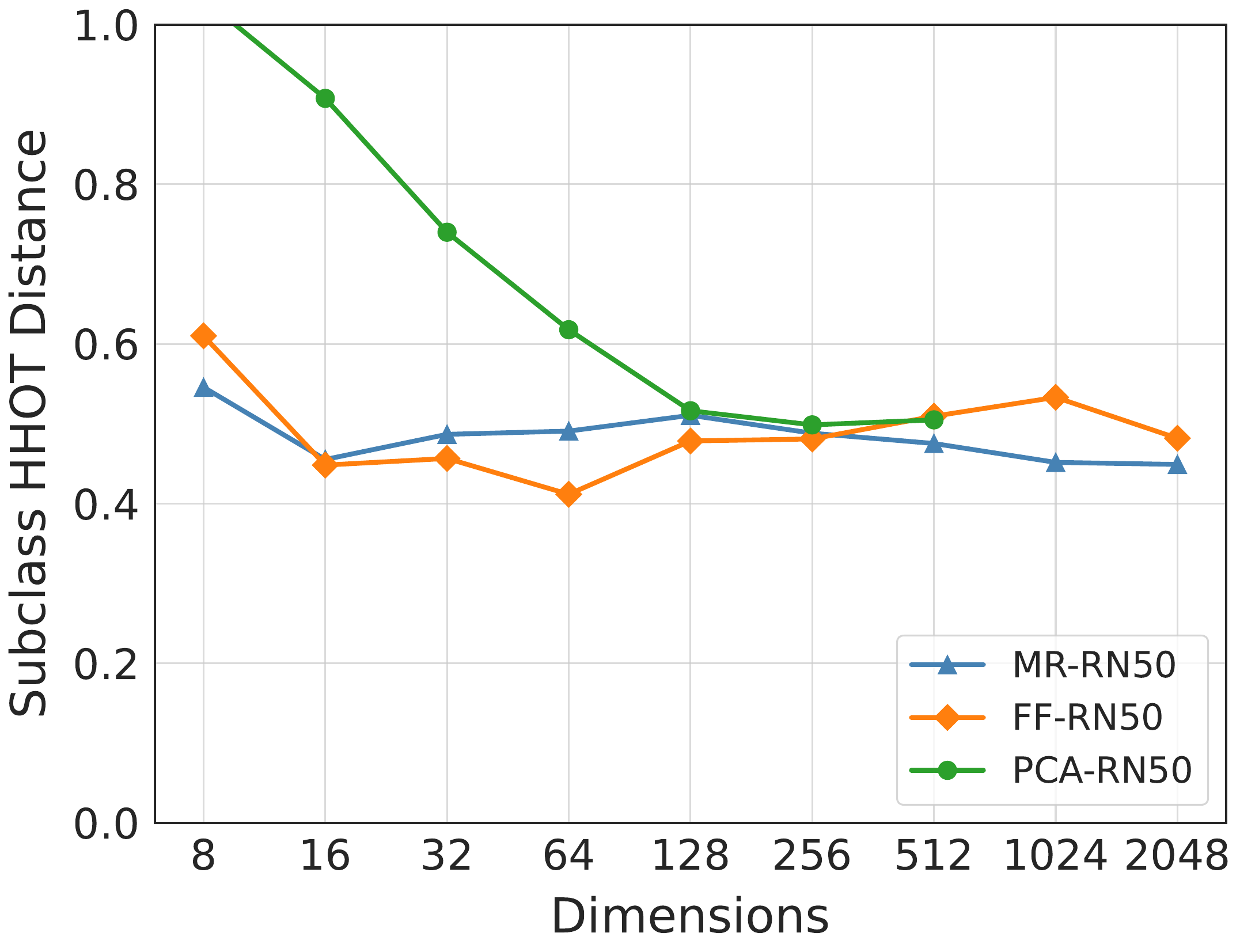}
      }
      \caption{MR struggles to outperform PCA embeddings for superclass HHOT distance. Still, as the number of dimensions increases MR easily beats out FF in superclass and subclass distances.}
      \label{fig:f3}
    \end{figure}

\subsection{High-Variance Dataset Clustering}
    \label{apdx:b3}
    \begin{figure}[h!]
       \centering

      {%
          \includegraphics[width=0.22\linewidth]{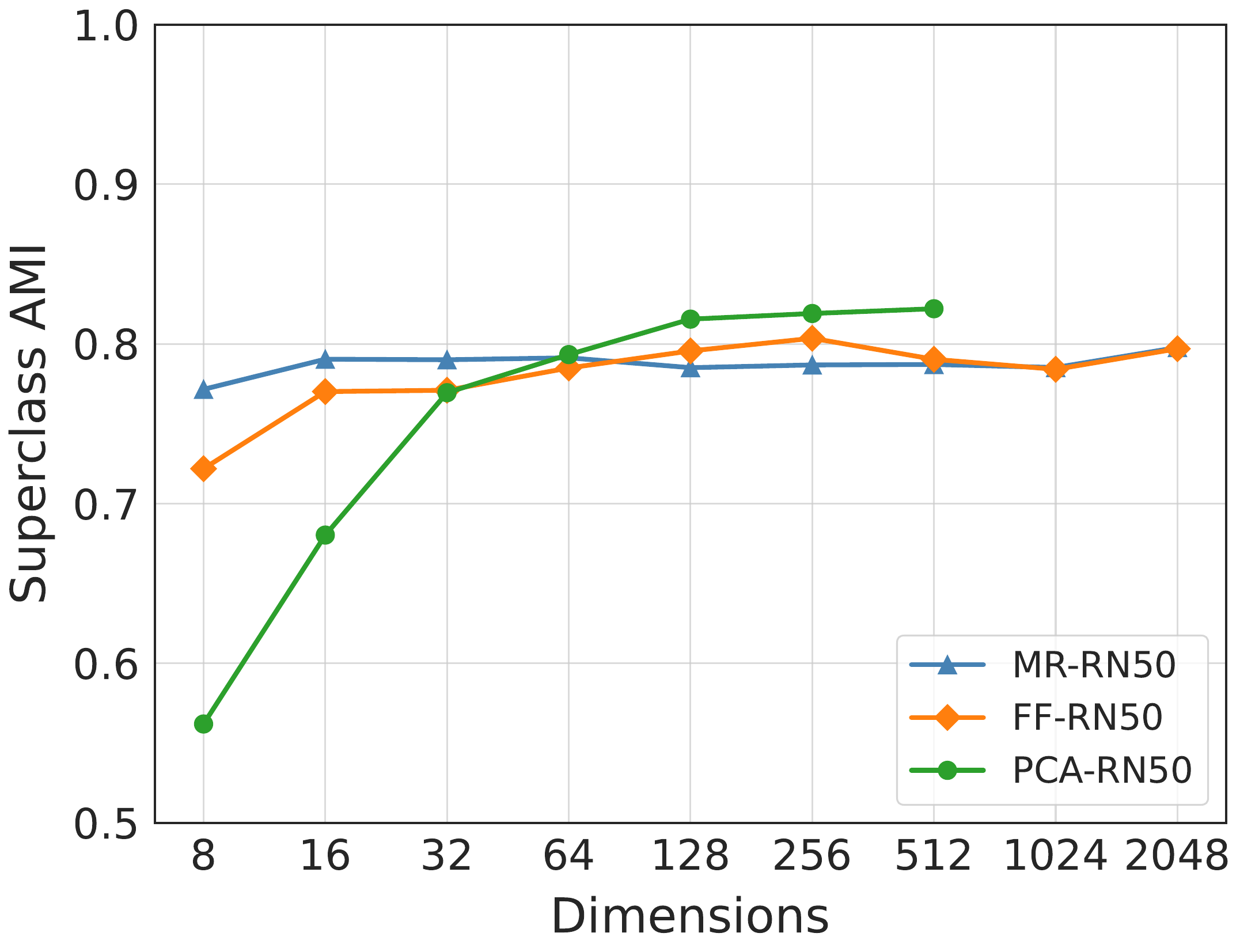}
          \includegraphics[width=0.22\linewidth]{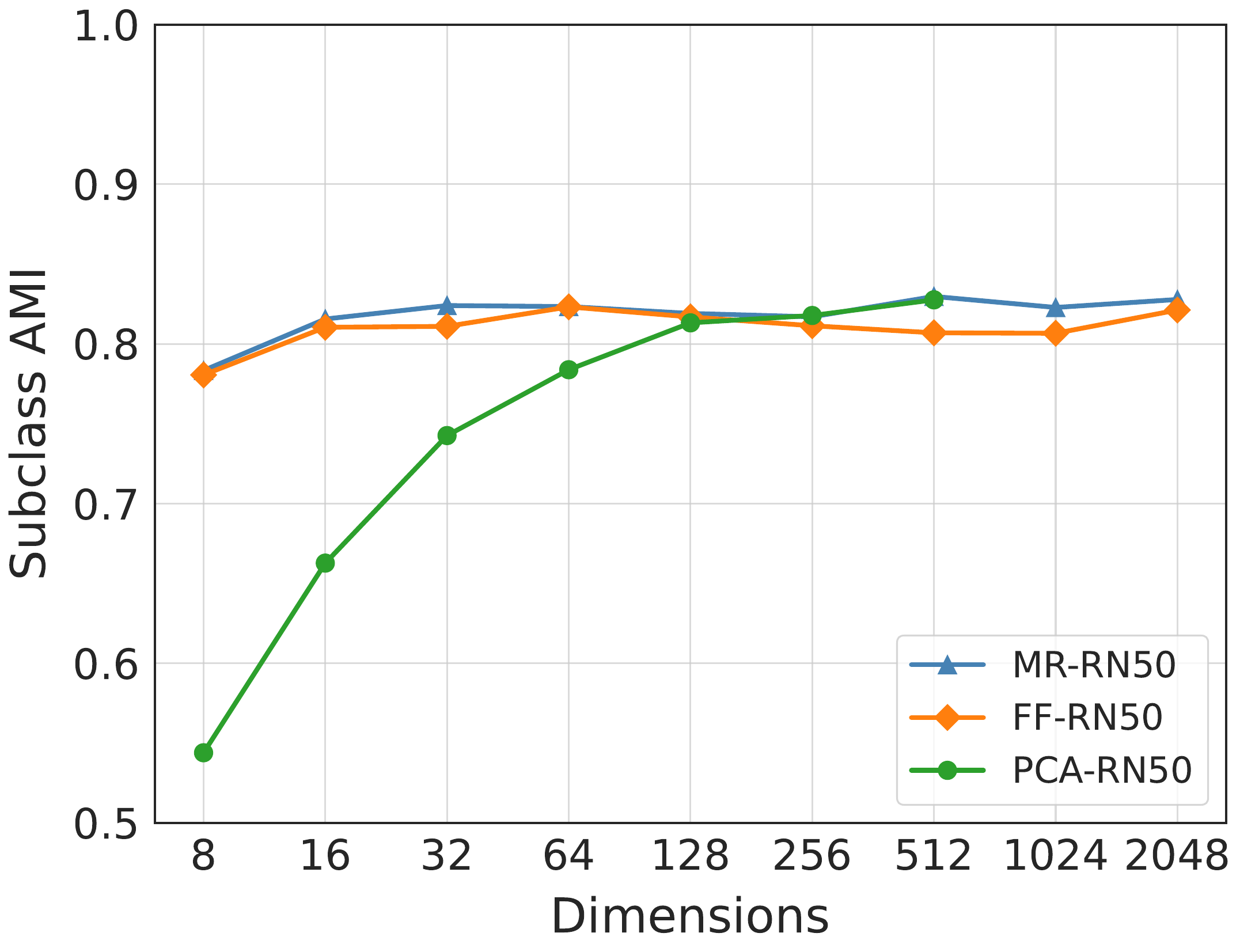}
          \includegraphics[width=0.22\linewidth]{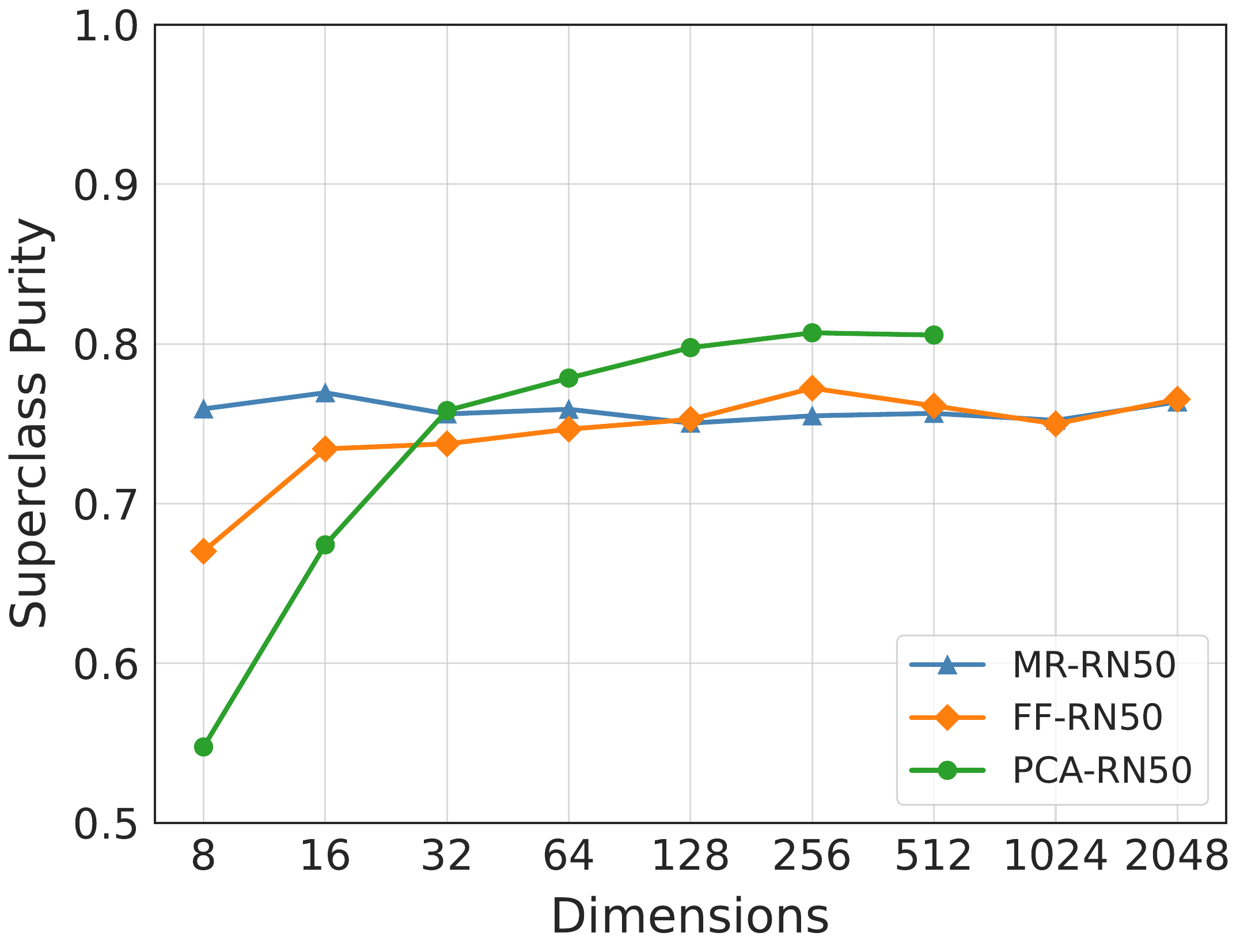}
          \includegraphics[width=0.22\linewidth]{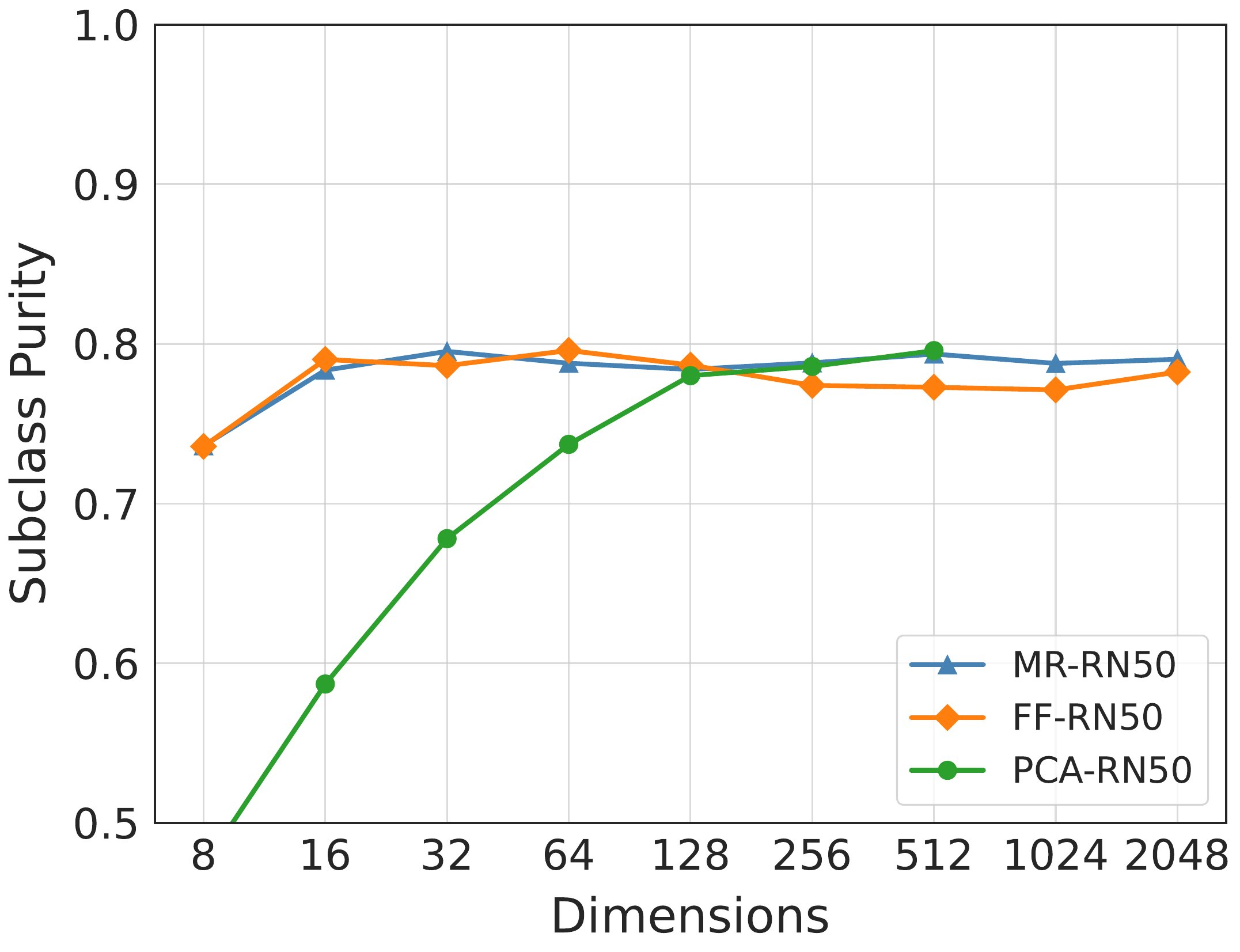}
      }
       \caption{When clustering on high-variance datasets, we find that while MR outperforms FF embeddings on subclass clustering, it is not a better option than PCA past 128 dimensions. Even more, FF and PCA embeddings outperform MR in both superclass AMI and superclass purity at high dimensions.}
      \label{fig:h2}
    \end{figure}
    \begin{figure}[htbp]
       \centering

      {%
          \includegraphics[width=0.4\linewidth]{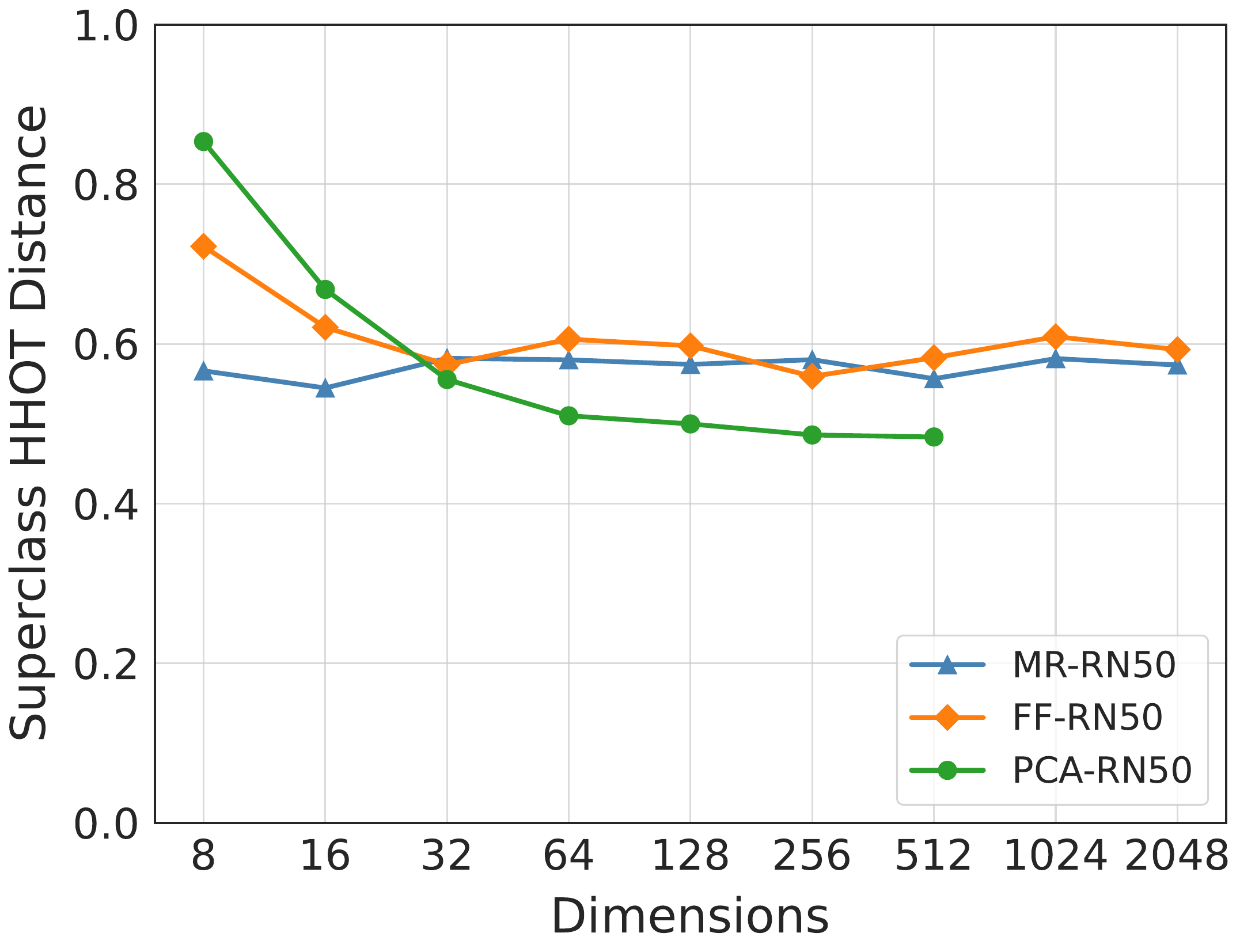}
          \includegraphics[width=0.4\linewidth]{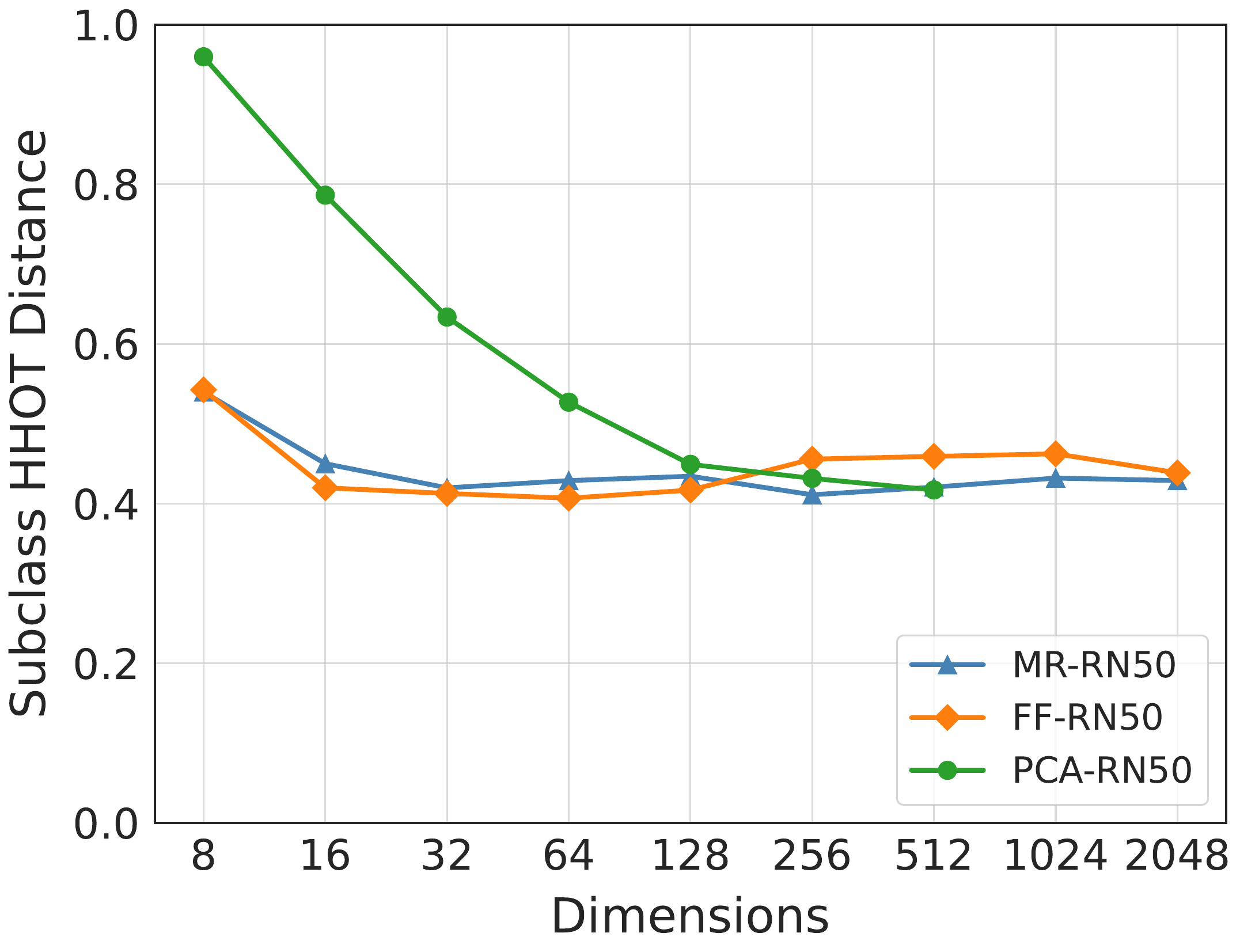}
      }
      \caption{MR's subclass clusters have the best HHOT distance of all embeddings. However, the HHOT distance of MR's superclass clustering is comparable to FF embeddings and definitively worse than PCA's.}
      \label{fig:h3}
    \end{figure}


\end{document}